\documentclass[nonacm,sigconf,natbib=true,anonymous=false]{acmart}
\AtBeginDocument{%
  \providecommand\BibTeX{{%
    \normalfont B\kern-0.5em{\scshape i\kern-0.25em b}\kern-0.8em\TeX}}}

\usepackage{xcolor}
\definecolor{darkgreen}{RGB}{84,130,53}

\usepackage{fancyvrb}
\usepackage{listings}
\usepackage{enumitem}
\usepackage{multirow}
\usepackage{stfloats}
\usepackage{subcaption}
\newcommand{\fig}[1]{Fig.~\ref{#1}}

\newcommand{\tb}[1]{Tab.~\ref{#1}}
\newcommand{\se}[1]{Section~\ref{#1}}

\usepackage{amsmath,amsfonts,bm}

\def\eqref#1{equation~\ref{#1}}

\def\1{\bm{1}}

\DeclareMathAlphabet{\mathsfit}{\encodingdefault}{\sfdefault}{m}{sl}
\SetMathAlphabet{\mathsfit}{bold}{\encodingdefault}{\sfdefault}{bx}{n}

\def\gM{{\mathcal{M}}}

\newcommand{\minisection}[1]{\vspace{3pt}\noindent\textbf{#1}}
\setlist[itemize]{leftmargin=10pt}

\begin{document}

\title{TRAD: Enhancing LLM Agents with Step-Wise Thought Retrieval and Aligned Decision}

\author{Ruiwen Zhou}
\email{skyriver@sjtu.edu.cn}
\affiliation{%
  \institution{Shanghai Jiao Tong University}
  \city{Shanghai}
  \country{China}
}

\author{Yingxuan Yang}
\email{zoeyyx@sjtu.edu.cn}
\affiliation{%
  \institution{Shanghai Jiao Tong University}
  \city{Shanghai}
  \country{China}
}

\author{Muning Wen}
\email{muningwen@sjtu.edu.cn}
\affiliation{%
  \institution{Shanghai Jiao Tong University}
  \city{Shanghai}
  \country{China}
}

\author{Ying Wen}
\email{ying.wen@sjtu.edu.cn}
\affiliation{%
  \institution{Shanghai Jiao Tong University}
  \city{Shanghai}
  \country{China}
}

\author{Wenhao Wang}
\email{wangwenhao-009@cpic.com.cn
}
\affiliation{%
  \institution{China Pacific Insurance}
  \city{Shanghai}
  \country{China}
}

\author{Chunling Xi}
\email{xichunling@cpic.com.cn}
\affiliation{%
  \institution{China Pacific Insurance}
  \city{Shanghai}
  \country{China}
}

\author{Guoqiang Xu}
\email{xuguoqiang-009@cpic.com.cn}
\affiliation{%
  \institution{China Pacific Insurance}
  \city{Shanghai}
  \country{China}
}

\author{Yong Yu}
\email{yyu@apex.sjtu.edu.cn}
\affiliation{%
  \institution{Shanghai Jiao Tong University}
  \city{Shanghai}
  \country{China}
}

\author{Weinan Zhang}
\email{wnzhang@sjtu.edu.cn}
\affiliation{%
  \institution{Shanghai Jiao Tong University}
  \city{Shanghai}
  \country{China}
}

\renewcommand{\shortauthors}{Zhou and Yang, et al.}

\begin{abstract}
    Numerous large language model (LLM) agents have been built for different tasks like web navigation and online shopping due to LLM's wide knowledge and text-understanding ability. Among these works, many of them utilize in-context examples to achieve generalization without the need for fine-tuning, while few of them have considered the problem of how to select and effectively utilize these examples. Recently, methods based on trajectory-level retrieval with task meta-data and using trajectories as in-context examples have been proposed to improve the agent's overall performance in some sequential decision making tasks. However, these methods can be problematic due to plausible examples retrieved without task-specific state transition dynamics and long input with plenty of irrelevant context. In this paper, we propose a novel framework (\emph{TRAD}) to address these issues. \emph{TRAD} first conducts \emph{\textbf{T}hought \textbf{R}etrieval}, achieving step-level demonstration selection via thought matching, leading to more helpful demonstrations and less irrelevant input noise. Then, \emph{TRAD} introduces \emph{\textbf{A}ligned \textbf{D}ecision}, complementing retrieved demonstration steps with their previous or subsequent steps, which enables tolerance for imperfect thought and provides a choice for balance between more context and less noise. Extensive experiments on ALFWorld and Mind2Web benchmarks show that TRAD not only outperforms state-of-the-art models but also effectively helps in reducing noise and promoting generalization. Furthermore, TRAD has been deployed in real-world scenarios of a global business insurance company and improves the success rate of robotic process automation. Our codes are available at: \url{https://github.com/skyriver-2000/TRAD-Official}.
\end{abstract}



\settopmatter{printacmref=false}

\maketitle

\section{Introduction}\label{sec:intro}

\begin{figure*}[tbp]
  \centering
  \vspace{-6pt}
  \includegraphics[width=0.98\linewidth, height=0.57\linewidth]{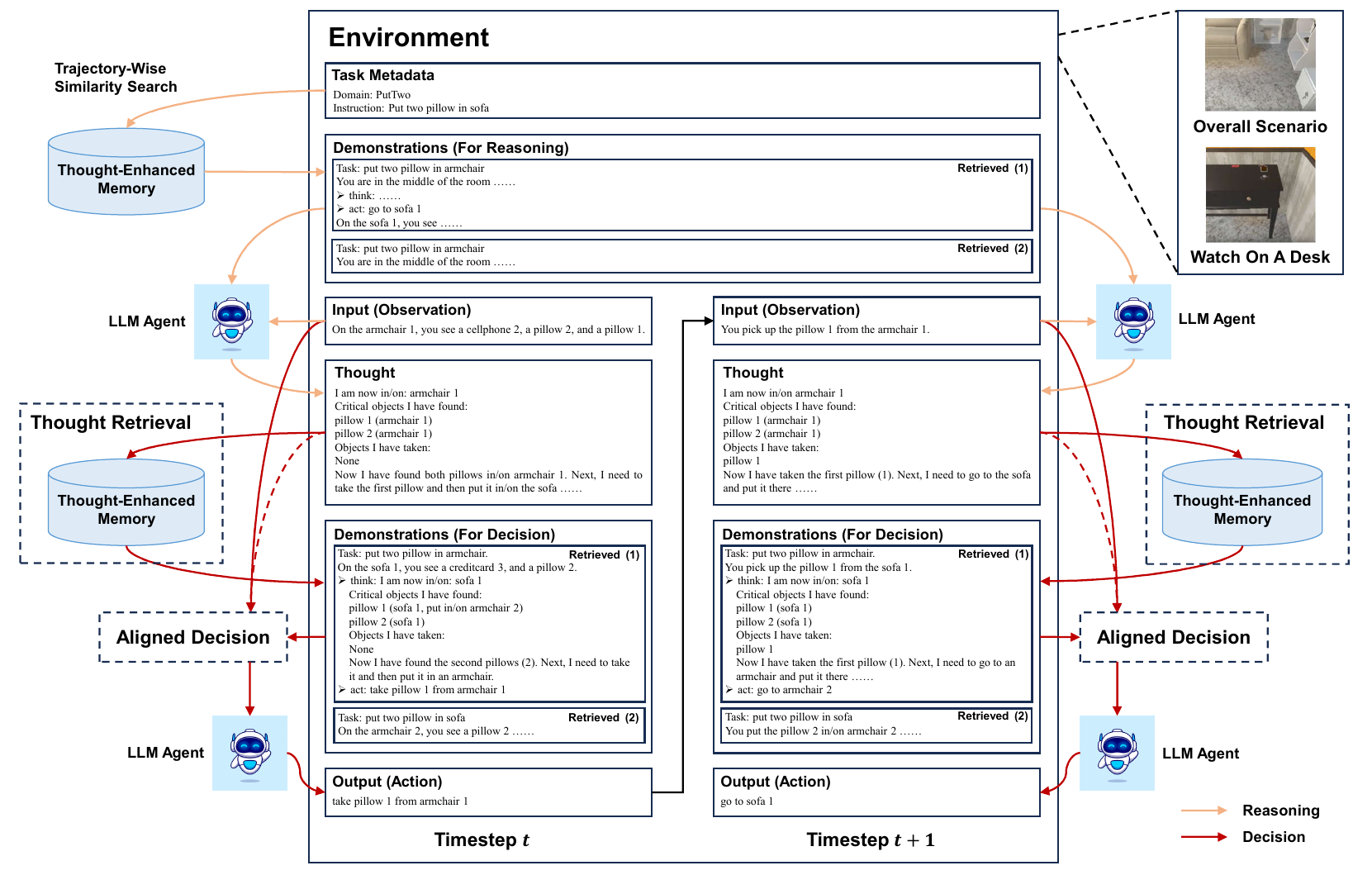}
  \vspace{-4pt}
  \caption{An overall illustration of \emph{TRAD} agent (on ALFWorld \cite{shridhar2021alfworld} enviroment). \emph{TRAD} first pre-processes expert trajectories, labeling each step with high-quality thoughts. At inference time, \emph{TRAD} first conducts \emph{thought retrieval}, which generates thought with trajectory-wise retrieved demonstrations as the query and keys for a more precise step-wise demonstration retrieval. Given the retrieved steps, TRAD employs \emph{aligned decision} module to complement their temporally neighboring steps and corresponding position information (\fig{fig:aligned-decision}). Finally, the next action is generated according to the enhanced demonstration.}
  \vspace{-4pt}
  \label{fig:trad-framework}
\end{figure*}

Large Language Models (LLMs) \cite{brown2020gpt3,touvron2023llama} have achieved remarkable success on various tasks like question answering \cite{zheng2023stepback}, chatbot \cite{ouyang2022training}, code synthesis \cite{roziere2023codellama}, text ranking \cite{ferraretto2023exaranker}, table-based reasoning \cite{ye2023dater}, and retrieval query expansion \cite{mackie2023grf} due to their wide knowledge and excellent ability of text understanding and generation. Recently, a series of works have attempted to build powerful agents based on LLMs for various sequential decision-making tasks, including text-based games \cite{yao2023tot}, online shopping \cite{yao2022webshop}, web navigation \cite{deng2023mind2web}, and information retrieval \cite{zhu2023large}.

Among existing LLM agents, some are trained with large-scale expert data by supervised fine-tuning (SFT) \cite{nakano2021webgpt,gur2023html,gur2023webagent}, while some are tuning-free and utilize in-context learning (ICL) with few expert demonstration examples \cite{yao2023react,kim2023rci,wang2023voyager,zheng2023synapse}. In this paper, we focus the scope on tuning-free ICL methods, as they are highly cost-effective and can seamlessly generalize to different tasks using only a small amount of expert samples. Most existing ICL-based agents are prompted with expert trajectories carefully selected by human \cite{wei2022cot,yao2023react,shinn2023reflexion}, which work well when few expert trajectories are available. However, when we have access to a large dataset of expert trajectories or an expert policy, the automatic and personalized selection of expert trajectories for each task instruction becomes necessary, and can have an essential influence on task performance.

Recently, \citet{zheng2023synapse} study the problem of demonstration selection and propose \emph{Synapse}, which retrieves relevant expert trajectories by task meta-data, and then prompts LLMs with these retrieved trajectories. \emph{Synapse} performs well on computer control tasks (MiniWob++ \cite{shi2017miniwob}) and web navigation tasks (Mind2Web \cite{deng2023mind2web}). Nevertheless, retrieving and prompting with complete trajectories can be problematic in the following three aspects.

\minisection{Plausible examples.} Sometimes generalization to data from various domains can be critical. For example, in cross-website and cross-domain subsets of Mind2Web, agents operate on websites unseen in the training set, i.e., memory. In this case, retrieving trajectories with only task meta-data is very likely to provide plausible examples, which share similar task instructions to the current one but require totally different solutions. As shown by experiments in \cite{zheng2023synapse}, plausible examples provide no more information than random examples and can usually mislead LLM agents to wrong decisions.

\minisection{Context limit of LLMs.} When facing tasks with long horizons and complex observations, prompting with complete trajectories will result in input sequences longer than the allowed length of LLMs. \emph{Synapse} thus has to reduce the number of trajectory examples or even fail to complete the task directly. Though some long-context LLMs can receive very long prompts, the performance can be harmed due to the issue of long-term forgetting \cite{lmsys2023longchat}.

\minisection{Irrelevant information in prompts.} 
LLMs are found sensitive to their prompts, and can easily copy their recent input \cite{radford2019gpt2,holtzman2020curious}. The decision at the current timestep can be related to very few steps in a retrieved trajectory, while other steps do not provide any helpful information. Therefore, irrelevant steps will have unpredictable effects on the decision of LLM agents. As shown by our experiments, they negatively impact the performance most of the time.

To address the problems of trajectory-wise retrieval and prompting, we delve into step-wise demonstration retrieval and prompting. We discover that, via demonstrating with relevant steps, the input context of the LLM agent can be significantly reduced. Thus, the issue of context limit and irrelevant information can be alleviated. Therefore, the critical part is to retrieve step demonstrations that are truly relevant and helpful. To achieve this, we utilize step-by-step reasoning, i.e. \emph{Chain-of-Thought} technique \cite{wei2022cot}, to abstract the state at each timestep as retrieval queries and keys. The generated \emph{thoughts} can involve historical information or future plans, which is more specific with state transitions and helpful in reducing plausible examples.

In this paper, we propose \emph{\textbf{T}hought \textbf{R}etrieval} and \emph{\textbf{A}ligned \textbf{D}ecision} (\emph{TRAD}), a novel framework that achieves step-wise demonstration retrieval via thought matching and enhances the context for action prediction with temporally neighboring steps and their order information. Our contribution can be summarized in four-folds:
\begin{itemize}
    \item We propose a \emph{thought retrieval} method, where we label thoughts for expert demonstration steps in advance with an LLM, prompt LLM agents to reason at inference time, and achieve step-wise retrieval by a similarity search on thought. To the best of our knowledge, this is the first work that enables the LLM agent with thought retrieval techniques for sequential decision-making.
    \item Based on the thought retrieval operation, we further propose an \emph{aligned decision} method, where we supply the retrieved steps with their temporal neighbors to overcome imperfect thoughts and enhance task-relevant information.
    \item We conduct extensive experiments and analysis on Mind2Web \cite{deng2023mind2web} tasks and ALFWorld \cite{shridhar2021alfworld}, showing that TRAD achieves state-of-the-art (SoTA) performance compared to existing works. \emph{TRAD} brings a 2.99\% improvement over the strongest baseline (93.78\% $\rightarrow$ 96.77\%) to the success rate (SR) on ALFWorld. On Mind2Web, \emph{TRAD} improves element accuracy, step SR, and SR remarkably over the powerful \emph{Synapse} agent \cite{zheng2023synapse} by 2.1\%, 1.4\%, and 0.5\%.
    \item We have deployed TRAD to the real-world robotic process automation scenarios of a global business insurance company, where \emph{TRAD} enables the LLM agent to significantly improve the success rate in a bunch of practical tasks. In average, \emph{TRAD} raises step SR from 90.2\% to 98.1\% and SR from 65.0\% to 92.5\%.
\end{itemize}
\section{Related Work}

\subsection{LLM Agents}
In recent years, there has been a rapidly growing trend to utilize pre-trained LLMs as the central controller to obtain human-level decision-making capabilities \cite{wang2023survey}. Among these works: \citet{nakano2021webgpt} fine-tune the GPT-3 \cite{brown2020gpt3} model for question answering in a text-based web browsing environment. \citet{yao2022webshop} develop WebShop, a simulated e-commerce website environment, and fine-tune a BERT \cite{devlin2018bert} model with imitation learning and reinforcement learning. \citet{yao2023react} insert a reasoning section between observation input and action output, significantly improving the performance on ALFWorld \cite{shridhar2021alfworld} and WebShop \cite{yao2022webshop} tasks. \citet{shinn2023reflexion} further improve over \cite{yao2023react} via verbally reflecting on linguistic task feedback signals. \citet{schick2023toolformer} teach LLMs to use external tools via simple APIs in a self-supervised learning way. \citet{park2023generative} introduce \emph{Generative Agents}, extending LLMs with natural language memories and retrieving them dynamically to plan behavior. \citet{wang2023deps} propose \emph{DEPS}, an interactive planning approach, which facilitates better error correction by integrating a description of the plan execution process and an explanation of failure feedback. \citet{wang2023voyager} employ an exploration curriculum, a growing skill library, and a novel iterative prompting mechanism, leading to better proficiency in playing Minecraft. \citet{deng2023mind2web} construct the Mind2Web dataset from real-world webpages, which consists of three subsets requiring different degrees of generalization, and compare the performance of imitation learning and few-shot inference.

As can be seen above, most existing LLM agents focus on: 1) improving task performance by direct fine-tuning \cite{nakano2021webgpt,yao2022webshop,deng2023mind2web}; 2) enhancing planning or reasoning by explicitly prompting \cite{yao2023react,shinn2023reflexion,wang2023deps}; 3) extending the application with an external memory or tool library \cite{schick2023toolformer,park2023generative,wang2023voyager}. However, providing more relevant information in prompts, as a fundamental way to elicit better task understanding, does not receive sufficient attention. When near-optimal demonstrations are accessible, selecting few-shot demonstrations properly can be a simple yet very effective way to improve task performance, which is investigated in our work.

\subsection{In-Context Example Selection}
LLMs have been shown excellence of few-shot learning \cite{brown2020gpt3}, and the selection of in-context examples can yield a significant improvement on the overall performance. \citet{liu2021good-example} first propose to retrieve the $k$-nearest neighbors ($k$-NN) of the input as in-context examples, and achieve improvement over random retrieval baselines. \citet{rubin2022learning} select relevant samples with an encoder trained with label similarity, and obtain better performance over BM25 and pre-trained encoder baselines. \citet{zhang2022unlabeled} consider selecting and labeling unlabeled examples as demonstrations to achieve the best performance, and view this problem as a sequential decision making task to solve by reinforcement learning. \citet{wu2023adaptive} further select examples in a subset recalled from $k$-NN search via minimizing the entropy of output.

\emph{IRCoT} \cite{trivedi2023ircot} should be the most relevant work to ours, which retrieves relevant documents with reasoning steps on question-answering tasks. However, their method consists of retrieving with a complete historical trajectory and accumulating retrieved trajectories over time, which are not transferable to complex sequential decision-making tasks, and we propose a method different from theirs in that: (i) Our method focuses on both providing more relevant demonstrations and reducing irrelevant context for sequential decision-making tasks, while theirs is limited to question-answering tasks and only addresses the first issue.
(ii) Our method retrieves completely different steps across timesteps and complements the retrieval results with temporal information, while theirs only accumulates relevant documents at every reasoning step and heuristically cuts off the earliest ones to fit in the context limit of LLMs.
(iii) Our method prepares pseudo-golden thoughts for expert trajectories in the memory to enable retrieval with trajectories without thoughts, and utilizes single-step thoughts as both queries and keys for precise retrieval, while theirs uses thoughts only as queries with raw documents as keys.

\begin{figure*}[tbp]
  \centering
  \vspace{-12pt}
  \includegraphics[width=0.98\linewidth]{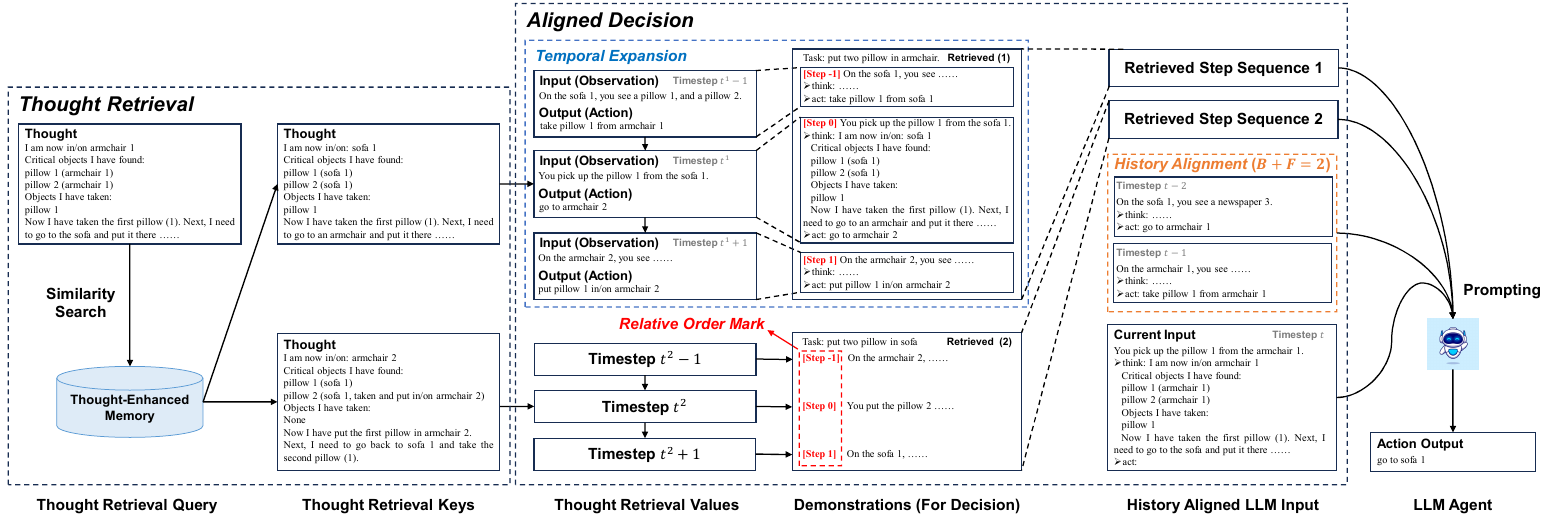}
  \vspace{-6pt}
  \caption{An illustration of our \emph{aligned decision} method, where $B=F=1$ and the $i$-th retrieved step is at time $t^i$ in its trajectory. The aligned decision method consists of three sub-processes to the retrieved step demonstrations and prompting: 1) \textcolor{blue}{Temporal Expansion:} Collect at most $B$ previous steps and $F$ subsequent steps for each retrieved step, and transform each step into a sequence of length $B+F+1$ from $t^i-B$ to $t^i+F$; 2) \textcolor{red}{Relative Order Mark:} For each step in one demonstration step sequence, we label its relative position to the retrieved step in this sequence, i.e., the previous one ($t^i-1$) with \texttt{[Step -1]} and the next one ($t^i+1$) with \texttt{[Step 1]}; 3) \textcolor{orange}{History Alignment:} For the current episode, we complement current observation (and thought, optional) with $B+F$ previous steps to enrich information and align with demonstrations.}
  \vspace{-4pt}
  \label{fig:aligned-decision}
\end{figure*}

The selection of in-context examples has been studied thoroughly for non-sequential tasks like question answering and sentiment analysis. However, for sequential decision-making tasks, how to select the examples to improve the overall performance remains unclear. \citet{zheng2023synapse} propose a trajectory-wise retrieval solution, while a more precise step-wise solution is still desired as discussed in \se{sec:intro}, which motivates our work.

\subsection{LLM Planning and Reasoning}\label{subsec:plan-reason}
Our work proposes to use thought, which can be viewed as a general abstraction of the current state, as queries and keys for retrieval. Nevertheless, plans, code comments, and any other text that extracts comprehensive information about the current state can serve as an alternative. Therefore, we particularly review some remarkable reasoning and planning works based on LLMs, and most of them are complementary to our work.

\citet{wei2022cot} first introduce the concept of \emph{Chain-of-Thought} (CoT) by providing with explicit step-by-step reasoning process in example outputs improving performance on arithmetic, commonsense, and symbolic reasoning tasks. \citet{wang2023sc} further find that a single reasoning path can be sub-optimal, and propose \emph{self-consistency} to address this problem by sampling multiple reasoning paths.
For efficient yet flexible search of reasoning paths, \citet{yao2023tot} apply tree search with self-evaluation to find globally excellent thoughts. \citet{besta2023got} later extend the tree-search structure to a graph search for even better flexibility and overall performance.

The works mentioned above consider problems that are non-sequential or solvable by a single complete reasoning path after receiving the input. For harder sequential decision-making problems: \citet{zhou2023least} introduce \emph{least-to-most} prompting to solve hard problems by decomposing the problem and solving sub-problems sequentially. \emph{ReAct} proposed by \citet{yao2023react} interacts with the environment in a reason-then-act style, which enriches the context for action prediction. \emph{Code-as-Policies} \cite{liang2023code} writes executable codes for embodied control by hierarchically expanding undefined programs, which can be viewed as implicit reasoning or CoT process. \citet{liu2023llm+p} propose to incorporate the strength of classical planners by translating the original problem into a PDDL \cite{aeronautiques1998pddl} problem to solve by classical planners. \citet{hao2023rap} and \citet{ding2023xot} share a similar insight that reasoning can be implemented indeed by planning, where \cite{hao2023rap} use LLMs as world models and \cite{ding2023xot} conduct MCTS for thought generation with a light-weight extra network.

To summarize, LLM planning and reasoning have continuously received huge attention from researchers in recent years. This makes our work flexible and improvable with more powerful planning and reasoning methods in the future.
\section{The TRAD Framework}

As discussed in \se{sec:intro}, trajectory-wise retrieving and prompting lead to issues of plausible examples, LLM context limits, and irrelevant information. To resolve these issues, we propose a novel method called \emph{\textbf{T}hought \textbf{R}etrieval} and \emph{\textbf{A}ligned \textbf{D}ecision} (\emph{TRAD}), as illustrated in \fig{fig:trad-framework}. 
Our \emph{TRAD} agent utilizes thought, which is obtained by reasoning about its current state, to retrieve similar steps from expert trajectories, and is then complemented with steps temporally correlated to the retrieved ones and their temporal position information to predict the action. Formally, our \emph{TRAD} agent can be summarized in one equation:
\begin{equation*}
    \pi_{TRAD}(a_t|\xi, o_{0:t}, a_{0:t-1}) = {\rm LLM}({\rm AD}({\rm TR}(\tau_t, \gM), \xi, o_{0:t}, a_{0:t-1}))~,
\end{equation*}
where $\xi$ is the current task, $o_{0:t}$ and $a_{0:t-1}$ are historical observations and actions, $\tau_t$ is the thought generated by LLM about the current state, TR and AD denote our \emph{thought retrieval} and \emph{aligned decision} modules, and $\gM$ refers to the thought-enhanced memory. We will present each module of \emph{TRAD} in the following subsections.

\subsection{Thought Preparation}\label{subsec:prepare-thoughts}
Most expert trajectories, collected by either human or other expert agents, do not contain their reasoning process. Therefore, before we utilize thoughts for retrieval, we should prepare thoughts for each demonstration step in the memory. Specifically, we start from a small subset of expert demonstrations and provide thoughts written by human experts for each step in it. Given this small subset as few-shot examples in prompts, we can query LLMs to label thoughts for a large memory. Although ground-truth actions are not accessible at inference time, we can prompt LLMs with them to generate thoughts of higher quality. In this way, LLMs produce pseudo-golden thoughts consistent with expert actions, and we obtain a \emph{thought-enhanced memory} $\gM$ supporting both trajectory-wise retrieval with task meta-data and step-wise retrieval with thoughts.

\subsection{Thought Retrieval}\label{subsec:thought-retrieval}
Given pseudo-golden thoughts for all steps in the memory, which can serve as keys for step-wise similarity search, we now present our \emph{thought retrieval} method to select relevant demonstrations at inference time. To be specific, we first conduct trajectory-wise demonstration retrieval as in \cite{zheng2023synapse} for thought generation. With these trajectory demonstrations, at each timestep $t$ we prompt the LLM to generate a thought $\tau_t$ for step-wise retrieval. Note that this process does not directly effects decision-making, hence it can be further simplified if necessary and the issues mentioned in \se{sec:intro} will not impact the agent severely.

With the thought $\tau_t$, which can be viewed as an abstraction, about current state, we conduct dense retrieval to find relevant steps in the \emph{thought-enhance memory} $\gM$. Here any encoder pre-trained on a large corpus for retrieval, e.g., Sentence-BERT \cite{reimers2019sbert} and DPR \cite{karpukhin2020dpr}, can be utilized to encode the query thought and key thoughts into dense vectors. Using a cosine similarity between the query and keys, we then collect top-$K$ relevant steps that belong to mutually different trajectories and their corresponding task instructions.

\subsection{Aligned Decision}\label{subsec:aligned-decision}
Now we have relevant demonstration steps from \emph{thought retrieval}. However, the query thought can be imperfect due to the lack of expert action information at inference time. As we will show by ablation experiments in \se{subsec:abl-study}, directly using these steps to form single-step demonstrations does not provide satisfactory performance, which is similar to the plausible example issue of trajectory-wise retrieval. Therefore, we propose an \emph{aligned decision} method to incorporate more information during the decision-making process. \emph{Aligned decision} complements LLM agents with steps temporally correlated to the retrieved ones and their temporal position information. As illustrated in \fig{fig:aligned-decision}, the \emph{aligned decision} method can be decomposed into following three sub-processes.

\minisection{Temporal expansion.} For each retrieved step, we first expand it into a step sequence involving $B$ previous steps and $F$ subsequent steps. When the number of previous or subsequent steps is smaller than $B$ or $F$, we simply take all previous or subsequent steps. This transforms each retrieved step into at most $(B+1+F)$ temporally successive steps, allowing LLM agents to correct their imperfect thoughts by looking at more related steps at decision-making time.

\minisection{Relative order mark.} Given $K$ expanded step sequences by \emph{temporal expansion}, we insert a mark for each step (including the retrieved ones) indicating the relative position w.r.t. its corresponding retrieved step, and incorporate this rule of mark in the prompt for decision. For example, the last step before the retrieved one will be marked as \texttt{[Step -1]}, the retrieved step as \texttt{[Step 0]}, and the first step after the retrieved one as \texttt{[Step 1]}. This provides temporal information about the $(B+1+F)\times K$ demonstration steps, and promotes more accurate demonstration following.

\minisection{History alignment.} Sometimes the optimal policy to a task, like ALFWorld, can be history-dependent, hence using single-step input for action prediction is unreasonable. Since we aim to reduce input content for less forgetting and noise, we should neither use all historical observations and actions. Moreover, even if we include previous actions as auxiliary information, there exists a mismatch where expert demonstrations are given as sequences of length $B+1+F$ while current input is a single step. We thus propose to insert at most $B+F$ previous input-output pairs (i.e. $o_{t-(B+F):t-1},~a_{t-(B+F):t-1}$) before current input $o_t$, transforming current input into a similar sequence to demonstrations.

\begin{table*}[tbp]
  \caption{Success Rate of Different Methods on 6 Types of ALFWorld Tasks. We compare \emph{TRAD} with \emph{ReAct} \cite{yao2023react}, \emph{Synapse} \cite{zheng2023synapse}, and their strong combination. \emph{TRAD} significantly outperforms all baselines in terms of overall performance, achieves the best performance in 5 out of 6 types of task, and shows a decent performance on Heat task. The improvement of \emph{TRAD} over all baselines on overall performance is statistically significant (measured by student's t-test at $p<0.05$).}
  \label{tab:exp-alfworld}
  \vspace{-8pt}
  \resizebox{\linewidth}{!}{
  \begin{tabular}{lccccccc}
    \toprule
    \textbf{Method} & \textbf{Put} & \textbf{Examine} & \textbf{Clean} & \textbf{Heat} & \textbf{Cool} & \textbf{PutTwo} & \textbf{All} \\
    \midrule
    ReAct (Random) & 0.8472$\pm$0.0393 & 0.8333$\pm$0.0454 & 0.9570$\pm$0.0304 & 0.8841$\pm$0.0205 & 0.9841$\pm$0.0224 & 0.8431$\pm$0.0277 & 0.8980$\pm$0.0093 \\
    ReAct (Fixed) & 0.7778$\pm$0.0708 & \textbf{0.9630$\pm$0.0262} & 0.9032$\pm$0.0263 & \textbf{0.9275$\pm$0.0205} & \textbf{1.0000$\pm$0.0000} & 0.8824$\pm$0.0480 & 0.9055$\pm$0.0186 \\
    Synapse & 0.9444$\pm$0.0196 & 0.7037$\pm$0.0262 & 0.9355$\pm$0.0000 & 0.9130$\pm$0.0615 & \textbf{1.0000$\pm$0.0000} & 0.8039$\pm$0.0555 & 0.8955$\pm$0.0106 \\
    Synapse + ReAct & 0.9167$\pm$0.0340 & 0.9444$\pm$0.0454 & \textbf{1.0000$\pm$0.0000} & 0.9130$\pm$0.0000 & 0.9524$\pm$0.0000 & 0.8627$\pm$0.0555 & 0.9378$\pm$0.0035 \\
    TRAD (Ours) & \textbf{0.9583$\pm$0.0000} & \textbf{0.9630$\pm$0.0524} & \textbf{1.0000$\pm$0.0000} & 0.8986$\pm$0.0205 & \textbf{1.0000$\pm$0.0000} & \textbf{0.9804$\pm$0.0277} & \textbf{0.9677$\pm$0.0141} \\
    \bottomrule
  \end{tabular}
  }
\end{table*}
\section{Experiments}

In this section, we aim to study the following research questions:
\begin{itemize}[leftmargin=22pt]
    \item [\textbf{RQ1}] How does \emph{TRAD} perform against existing SoTA methods?
    \item [\textbf{RQ2}] Does \emph{thought retrieval} help to reduce irrelevant context and improve the overall performance?
    \item [\textbf{RQ3}] Does \emph{aligned decision} help to supply information when generalization is important?
    \item [\textbf{RQ4}] Diving into \emph{aligned decision}, are all \emph{temporal expansion} (TE), \emph{relative order mark} (ROM), and \emph{history alignment} (HA) necessary for improvement?
    \item [\textbf{RQ5}] How will the performance and advantage of \emph{TRAD} be effected by critical hyper-parameters?
\end{itemize}

\subsection{Experiment Setup}
To answer the above research questions, we conduct extensive experiments on ALFWorld \cite{shridhar2021alfworld} and Mind2Web \cite{deng2023mind2web} tasks. For each task, we introduce the details of evaluation as follows.

\vspace{5pt}\noindent\textbf{ALFWorld} \cite{shridhar2021alfworld} is a text-based game aligned with ALFRED \cite{shridhar2020alfred} benchmark. It involves 6 types of tasks where an agent must take a series of actions (e.g. \emph{go to shelf 1}, \emph{take vase 2 from shelf 1}, \emph{put vase 2 in/on cabinet 5}) to achieve a high-level goal given by a natural language instruction (e.g. \emph{put some vase on a cabinet}). This environment is challenging in three aspects: 1) Agent should determine likely places of a householding object and explore them one by one to find such object; 2) Agent should understand the usage of some objects like microwaves, fridges, and desklamps; 3) Some tasks can take an agent more than 30 steps to solve, requiring substantial long-term memorization.

Following \citet{shridhar2021alfworld}, we evaluate on the subset of 134 out-of-distribution tasks, comparing the task success rates of \emph{TRAD} to \emph{ReAct} \cite{yao2023react} and \emph{Synapse} \cite{zheng2023synapse} (without state abstraction as observations are short). As \emph{ReAct} and \emph{Synapse} has provided sufficiently strong performances, we do not include more complex reasoning and planning baselines and corresponding variants of \emph{TRAD} due to our API cost limit.
Note that the original \emph{ReAct} uses fixed but not retrieved trajectories as demonstrations, hence we test two \emph{ReAct} baselines to eliminate such an effect:
\begin{itemize}
    \item \emph{ReAct} (Fixed) uses fixed human-written trajectories as demonstrations;
    \item \emph{ReAct} (Random) randomly samples trajectories from the memory as demonstrations.
\end{itemize}

For fair comparison, \emph{TRAD} uses thoughts in exactly the same format as \emph{ReAct}, and shares a consistent memory of expert trajectories with \emph{Synapse}. We also add a strong baseline (\emph{Synapse}+\emph{ReAct}) combining the trajectory-level retrieval in \emph{Synapse} and the reasoning in \emph{ReAct}. On ALFWorld, all methods are built with GPT-4 \cite{openai2023gpt4} and 2 in-context examples.

\vspace{5pt}\noindent\textbf{Mind2Web} \cite{deng2023mind2web} is an HTML-based web navigation benchmark collected from real-world webpages, involving various tasks such as searching, trip booking, social network subscription, etc. It contains 3 subsets, i.e., cross-task, cross-website, cross-domain. This environment is challenging in two aspects: 1) Existing LLM agents can hardly understand HTML input well; 2) Unseen tasks and websites can require substantial generalization. \citet{deng2023mind2web} find that the cross-website and cross-domain subsets are significantly harder due to the need for generalization to unseen websites.

Since Mind2Web was introduced only about half a year ago, there is a lack of suitable baseline algorithms, and thus we compare our \emph{TRAD} agent to \emph{Synapse} \cite{zheng2023synapse} and \emph{ReAct} \cite{yao2023react}. Following \citet{zheng2023synapse}, we evaluate on all 3 subsets, comparing the element accuracy (Ele. Acc), step success rate (Step SR), and trajectory success rate (SR). For fair comparison, we follow \cite{zheng2023synapse} and summarize observations into 5 web elements with the pre-trained element ranker provided by \cite{deng2023mind2web} for all methods. Since the observations are still very complex on Mind2Web, including thoughts for every step in trajectories is not available, hence: 1) we do not include a \emph{Synapse} + \emph{ReAct} baseline; 2) \emph{TRAD} generates thoughts and predicts actions by a single-step prompt with the current observation and previous actions (without previous observations). To eliminate the effect of prompting style and reasoning, we build two \emph{ReAct} baselines using the same format of prompt as \emph{TRAD}:
\begin{itemize}
    \item \emph{ReAct} (Random), for which we prompt \emph{ReAct} with completely random demonstration steps.
    \item \emph{ReAct} (Relevant), for which we prompt \emph{ReAct} with demonstrate steps randomly chosen from trajectories retrieved by \emph{Synapse}.
\end{itemize}
We do not include the \emph{ReAct} (Fixed) baseline as it is hard to write or pick demonstrations commonly helpful for such diverse test sets. 
We also provide the results of the simplest MindAct \cite{deng2023mind2web} baseline without reasoning and retrieval for completeness. On Mind2Web, all methods are built with GPT-3.5-turbo and 3 in-context examples.

\begin{table*}[tbp]
    \caption{Results (\%) of all methods on Mind2Web benchmark. \emph{TRAD} achieves the best overall performances and the most improvement on the two harder subsets, especially the most out-of-distribution Cross-Domain subset. The improvement of \emph{TRAD} over all baselines on three overall metrics is statistically significant (measured by student's t-test with $p<0.01$).}
    \label{tab:exp-mind2web}
    \vspace{-8pt}
    \resizebox{\linewidth}{!}{
    \begin{tabular}{lccccccccccccccc}
    \toprule
    \multirow{2}{*}{\textbf{Method}} & \multicolumn{3}{c}{\textbf{Cross-Task}} & & \multicolumn{3}{c}{\textbf{Cross-Website}} & & \multicolumn{3}{c}{\textbf{Cross-Domain}} & & \multicolumn{3}{c}{\textbf{All}} \\
    \cline{2-4} \cline{6-8} \cline{10-12} \cline{14-16}
    & \multicolumn{1}{c}{\textbf{Ele. Acc}} & \multicolumn{1}{c}{\textbf{Step SR}} & \multicolumn{1}{c}{\textbf{SR}} & & \multicolumn{1}{c}{\textbf{Ele. Acc}} & \multicolumn{1}{c}{\textbf{Step SR}} & \multicolumn{1}{c}{\textbf{SR}} & & \multicolumn{1}{c}{\textbf{Ele. Acc}} & \multicolumn{1}{c}{\textbf{Step SR}} & \multicolumn{1}{c}{\textbf{SR}} & & \multicolumn{1}{c}{\textbf{Ele. Acc}} & \multicolumn{1}{c}{\textbf{Step SR}} & \multicolumn{1}{c}{\textbf{SR}} \\
    \midrule
    MindAct & 20.3 & 17.4 & 0.8 & & 19.3 & 16.2 & 0.6 & & 21.0 & 18.6 & 1.0 & & 20.6 & 18.0 & 0.9 \\
    ReAct (Random) & 31.0 & 24.7 & 1.6 & & 25.7 & 19.1 & 0.6 & & 27.9 & 22.9 & 1.8 & & 28.3 & 22.7 & 1.6 \\
    ReAct (Relevant) & 31.3 & 26.0 & 1.2 & & 26.7 & 20.5 & 0.6 & & 28.0 & 23.1 & 1.6 & & 28.5 & 23.4 & 1.4 \\
    Synapse w/o Retrieval & 33.1 & 28.9 & 3.2 & & 27.8 & 22.1 & \textbf{1.1} & & 30.0 & 26.5 & 1.4 & & 30.4 & 26.4 & 1.7 \\ 
    Synapse & 34.4 & 30.6 & 2.0 & & 28.8 & 23.4 & \textbf{1.1} & & 29.4 & 25.9 & 1.6 & & 30.4 & 26.6 & 1.6 \\
    TRAD (Ours) & \textbf{35.2} & \textbf{30.8} & \textbf{3.6} & & \textbf{30.4} & \textbf{24.0} & 0.6 & & \textbf{32.0} & \textbf{28.0} & \textbf{2.0} & & \textbf{32.5} & \textbf{28.0} & \textbf{2.1} \\
    \bottomrule
    \end{tabular}
    }
\end{table*}

\subsection{Evaluation on ALFWorld}

The success rate of each method tested on ALFWorld is shown in \tb{tab:exp-alfworld}. Generally, our \emph{TRAD} agent achieves an average success rate of 96.77\%, significantly outperforming \emph{ReAct} ($\sim$90\%), \emph{Synapse} (89.55\%), and even their strong combination (93.78\%). It is also worth noting that the worst trial of \emph{TRAD} among 3 random seeds achieves a success rate of 94.8\%, outperforming the best trial produced by any other method (94.0\%).

Down to the success rate on each type of task, we observe that the success rate of each method varies more on the simplest \emph{Put} task and the hardest \emph{PutTwo} task. We discuss the results of these two tasks respectively as follows:
\begin{itemize}
    \item On the simplest \emph{Put} task, \emph{ReAct} performs even more poorly than other harder tasks. We find that the two vital reasons for \emph{ReAct}'s failure on \emph{Put} task are incorrect location and usage of objects, e.g. trying to put an object in a closed safe. As this issue can be alleviated through a combination with \emph{Synapse}, the necessity of retrieving relevant demonstrations thus justified.
    \item \emph{TRAD} achieves the largest improvement on the hardest \emph{PutTwo} task. \emph{PutTwo} requires to correct the locations of two objects and a comprehensive understanding of its task process. Since \emph{TRAD}'s outstanding performance on this hardest task is obtained from a reduced input context at decision-making time, we can conclude that step-wise \emph{thought retrieval} is helpful by reducing the noise of irrelevant steps and finding relevant examples more precisely.
\end{itemize}

\subsection{Evaluation on Mind2Web}

To verify the capability of \emph{TRAD} under more realistic scenarios, we compare \emph{TRAD} to \emph{ReAct} and the current SoTA method, \emph{Synapse}, on the Mind2Web benchmark, and the results are shown in \tb{tab:exp-mind2web}. We also include the results of \emph{Synapse} without retrieval here to better illustrate the effect of different retrieval methods.

Generally, \emph{TRAD} achieves the highest performance in terms of all 3 metrics averaged on 3 subsets. Considering that the trajectory-level retrieval of \emph{Synapse} only brings marginal boosts on Cross-Task and Cross-Website subsets, and even slightly impacts the performance on the Cross-Domain subset, our \emph{TRAD} method can be thus justified in two aspects:
\begin{itemize}
    \item By reducing input context and utilizing step-wise relevant demonstrations, our step-wise \emph{thought retrieval} helps more than the trajectory-wise retrieval with task meta-data in \emph{Synapse} to improve on the simplest Cross-Task subset.
    \item By eliminating plausible examples and complementing temporal correlated steps, \emph{aligned decision} helps to improve on the two harder subsets, especially the most out-of-distribution Cross-Domain subset.
\end{itemize}

Furthermore, we observe that the two \emph{ReAct} baselines perform poorly on this task, which indicates that:
\begin{itemize}
    \item The thoughts generated by GPT-3.5-turbo on Mind2Web tasks are not sufficient for LLM agents to infer the correct action.
    \item The single-step prompting style which removes previous observations does not benefit overall performance.
\end{itemize}
On the contrary, \emph{TRAD} utilizes these imperfect thoughts for retrieval rather than direct decision-making, and is complemented with temporally correlated steps via \emph{aligned decision}. Therefore, \emph{TRAD} is not negatively impacted by the imperfect thoughts, but transforms them into helpful information.

Before we start the study on detailed design and hyper-parameter choices of \emph{TRAD}, we can summarize our performance evaluation on ALFWorld and Mind2Web benchmarks and answer the first three research questions as follows.

\minisection{Answer to RQ1:} On both householding (ALFWorld) and web navigation (Mind2Web) tasks, \emph{TRAD} significantly outperforms curernt SoTA methods and becomes the new SoTA method.

\minisection{Answer to RQ2:} On ALFWorld benchmark, \emph{Synapse} + \emph{ReAct} generates thoughts in exactly the same way with our \emph{TRAD}, and uses entire relevant trajectories (more information than \emph{TRAD}) as demonstrations for action prediction. However, \emph{TRAD} shows obvious advantage over this baseline. Therefore, we can conclude that \emph{TRAD} benefits from more relevant demonstrations and less irrelevant input context brought by \emph{thought retrieval}.

\minisection{Answer to RQ3:} On Mind2Web benchmark, \emph{TRAD} achieves the most improvement over \emph{Synapse} on the Cross-Domain subset which requires the most generalization. Therefore, we can tell that the \emph{aligned decision} method complements critical information for decision-making on unseen input.

\subsection{Ablation Studies}\label{subsec:abl-study}
We have verified the effectiveness of \emph{TRAD} on two different scenarios, i.e., automatic householding and web navigation. Next, we are to examine the effect of each module in \emph{TRAD}. Due to our limited budget for API usage, all ablation studies are conducted on the Mind2Web benchmark with GPT-3.5-turbo.

\subsubsection{The Effect of Aligned Decision}
First, we study the effect of macro building blocks of \emph{TRAD}. Since eliminating \emph{thought retrieval} will disable \emph{aligned decision} at the same time and break the framework fundamentally, we do not remove the \emph{thought retrieval} module, but ablate each component of \emph{aligned decision}, i.e., \emph{temporal expansion} (TE), \emph{relative order mark} (ROM), and \emph{history alignment} (HA), and compare the corresponding performances. The results are shown in \tb{tab:exp-mind2web-abl}.

\begin{table*}[tbp]
    \caption{Results (\%) of ablation studies on Mind2Web benchmark. TE builds the basic structure of \emph{aligned decision} and is thus critical for performance boost on all three subsets. HA and ROM work well to promote generalization on the two harder Cross-Website and Cross-Domain subsets but provide little help on the Cross-Task subset. The improvement of \emph{TRAD} over all ablation baselines on Ele. Acc and Step SR is statistically significant (measured by student's t-test with $p<0.05$).}
    \label{tab:exp-mind2web-abl}
    \vspace{-8pt}
    \resizebox{\linewidth}{!}{
    \begin{tabular}{lccccccccccccccc}
    \toprule
    \multirow{2}{*}{\textbf{Method}} & \multicolumn{3}{c}{\textbf{Cross-Task}} & & \multicolumn{3}{c}{\textbf{Cross-Website}} & & \multicolumn{3}{c}{\textbf{Cross-Domain}} & & \multicolumn{3}{c}{\textbf{All}} \\
    \cline{2-4} \cline{6-8} \cline{10-12} \cline{14-16}
    & \multicolumn{1}{c}{\textbf{Ele. Acc}} & \multicolumn{1}{c}{\textbf{Step SR}} & \multicolumn{1}{c}{\textbf{SR}} & & \multicolumn{1}{c}{\textbf{Ele. Acc}} & \multicolumn{1}{c}{\textbf{Step SR}} & \multicolumn{1}{c}{\textbf{SR}} & & \multicolumn{1}{c}{\textbf{Ele. Acc}} & \multicolumn{1}{c}{\textbf{Step SR}} & \multicolumn{1}{c}{\textbf{SR}} & & \multicolumn{1}{c}{\textbf{Ele. Acc}} & \multicolumn{1}{c}{\textbf{Step SR}} & \multicolumn{1}{c}{\textbf{SR}} \\
    \midrule
    TRAD w/o TE & 34.2 & 28.4 & 1.2 & & 27.4 & 20.4 & 0.6 & & 29.1 & 24.0 & 1.4 & & 30.0 & 24.5 & 1.3 \\
    TRAD w/o HA & \textbf{36.2} & \textbf{31.1} & \textbf{4.0} & & 28.3 & 22.2 & 0.6 & & 29.4 & 24.9 & 1.8 & & 30.8 & 25.9 & \textbf{2.1} \\ 
    TRAD w/o ROM & 35.7 & 30.5 & 3.6 & & 28.9 & 22.3 & 0.6 & & 31.5 & 27.2 & 1.9 & & 32.1 & 27.2 & 2.0 \\
    TRAD (Ours) & 35.2 & 30.8 & 3.6 & & \textbf{30.4} & \textbf{24.0} & 0.6 & & \textbf{32.0} & \textbf{28.0} & \textbf{2.0} & & \textbf{32.5} & \textbf{28.0} & \textbf{2.1} \\
    \bottomrule
    \end{tabular}
    }
\end{table*}

From \tb{tab:exp-mind2web-abl}, we observe that the performance without each component varies differently on the simplest Cross-Task subset and the two harder subsets:
\begin{itemize}
    \item On the harder Cross-Website and Cross-Domain subsets, the elimination of all three modules in \emph{aligned decision} results in a significant performance drop, and the effect of \emph{temporal expansion} is the most significant. This is intuitive, since only retrieved steps are provided to the agent without TE, and thus the agent becomes more vulnerable to imperfect thoughts.
    \item On the simplest Cross-Task subset, however, \emph{history alignment} and \emph{relative order mark} are not that helpful and even cause performance drop. As discussed earlier (\se{sec:intro} and \se{subsec:aligned-decision}), when the issue of plausible examples is not severe, reducing context and prompting with the most relevant demonstration becomes the dominant factor of performance boost. Therefore, only \emph{temporal expansion} remains beneficial for recovering from imperfect thoughts, while the other two components lead to sub-optimal performance.
\end{itemize}

Generally, the \emph{aligned decision} method provides more information about the source trajectories of retrieved steps and the current trajectory, and helps especially for scenarios where generalization is essential. We can now summarize these observations and answer the fourth research question.

\minisection{Answer to RQ4:} Among the sub-processes in \emph{aligned decision}, 1) \emph{temporal expansion} provides tolerance for imperfect thoughts and improves the overall performance of \emph{TRAD} consistently; 2) \emph{relative order mark} and \emph{history alignment} complement \emph{TRAD} with temporal information about the trajectories of retrieved steps and the current trajectory, which serve as useful context for out-of-distribution decision-making but may become less useful for in-distribution decision-making.

\subsubsection{The Effect of Expansion Steps $B$ and $F$}\label{subsubsec:abl-temporal-extension}
\begin{figure}[htbp]
    \centering
    \begin{subfigure}[b]{0.22\textwidth}
        \includegraphics[width=\textwidth]{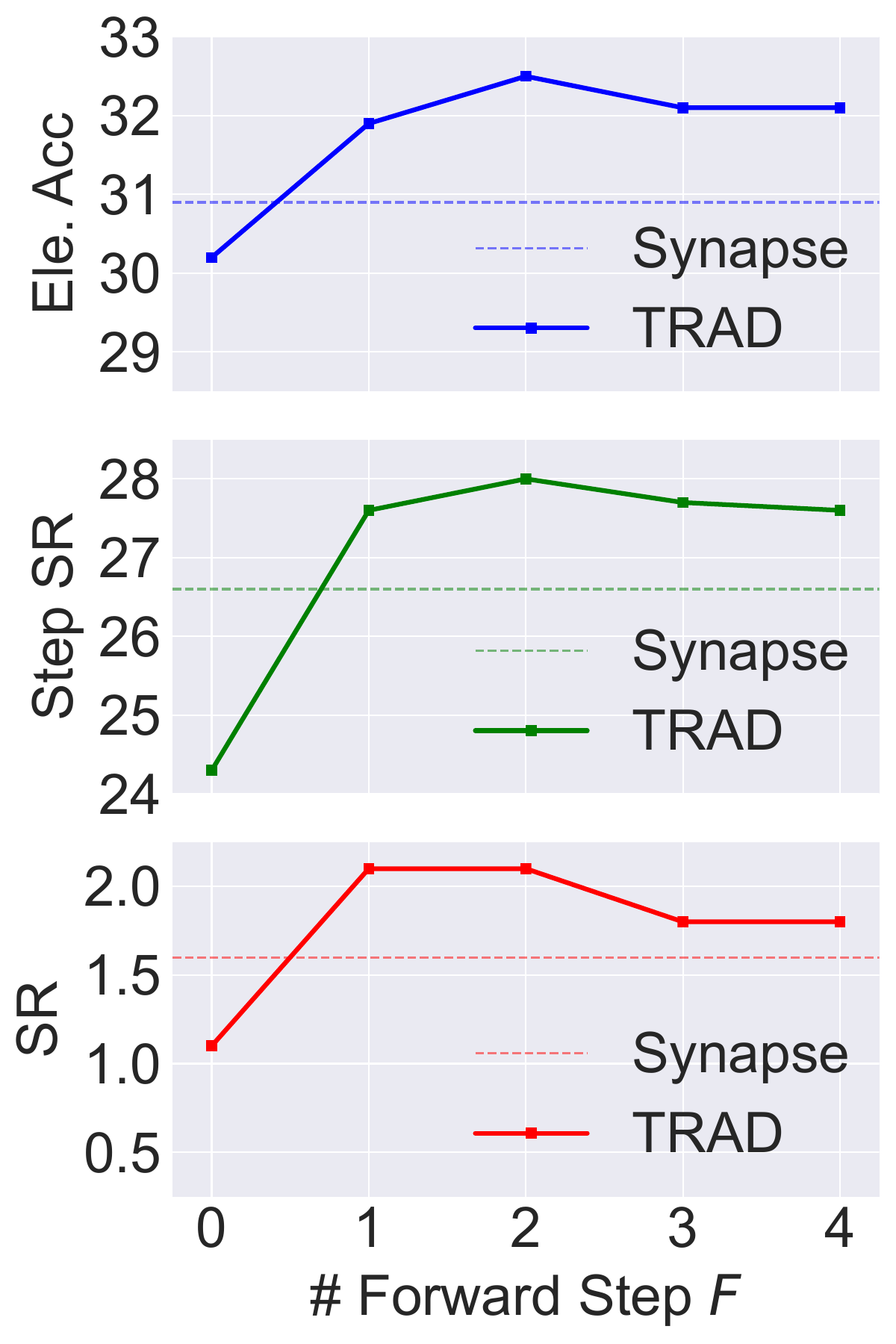}
        \vspace{-16pt}
        \caption{Varying $F$}
        \label{fig:average-forward-0}
    \end{subfigure}~~~~~~~~
    \begin{subfigure}[b]{0.22\textwidth}
        \includegraphics[width=\textwidth]{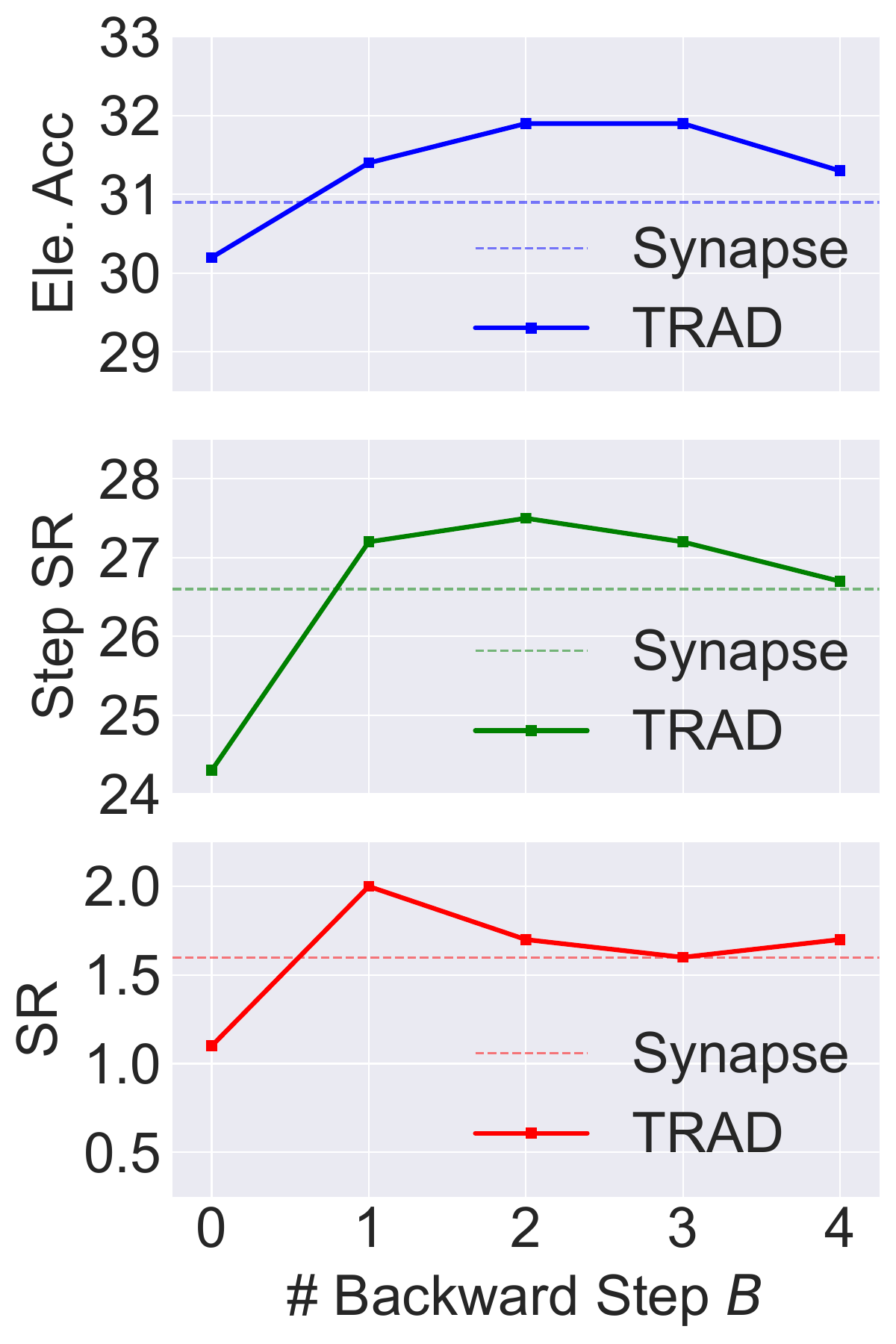}
        \vspace{-16pt}
        \caption{Varying $B$}
        \label{fig:average-backward-0}
    \end{subfigure}
    \vspace{-8pt}
    \caption{The effect of varying subsequent steps $F$ and previous steps $B$ on Mind2Web benchmark. Solid lines correspond to the performance metrics of \emph{TRAD} given different $F$ and $B$, and the dashed lines correspond to the \emph{Synapse} baseline. Forward expansion ($F>0$) generally provides more improvement than backward expansion ($B>0$) over no expansion ($F=B=0$) and the \emph{Synapse} baseline. $F$ or $B$ does not help more when they are sufficiently large.}
    \vspace{-8pt}
    \label{fig:exp-expansion-step}
\end{figure}

Next we vary a critical hyper-parameter, the number of temporal expansion steps, and investigate how the overall performance will change accordingly. To avoid an expensive grid search on $B$ and $F$, we consider only one-side expansion by varying $B$ or $F$ from $0$ to $4$ with the other set to $0$. The results over all 3 subsets are shown in \fig{fig:exp-expansion-step}. 

From \fig{fig:exp-expansion-step}, we can have the following observations:
\begin{itemize}
    \item Both forward expansion ($F>0$) and backward expansion ($B>0$) achieve improvement compared to no expansion ($F=B=0$). This justifies our design of \emph{aligned decision}.
    \item Either forward expansion or backward expansion does not benefit from increasing a large enough $F$ or $B$ further. This proves our hypothesis that irrelevant context too far from the current state is of little value and even noisy.
    \item Generally, forward expansion performs better than backward expansion when varying $F$ and $B$. The reason for this phenomenon might be that historical information has been incorporated in thoughts and thus future information helps more.
    \item \emph{TRAD} achieves its best performance when $F=2$ and $B=0$, and consistently outperforms \emph{Synapse} with forward expansion.
\end{itemize}

\subsubsection{The Effect of Demonstration Amount $K$}\label{subsubsec:abl-k}
\begin{figure}[htbp]
    \centering
    \includegraphics[width=0.96\columnwidth]{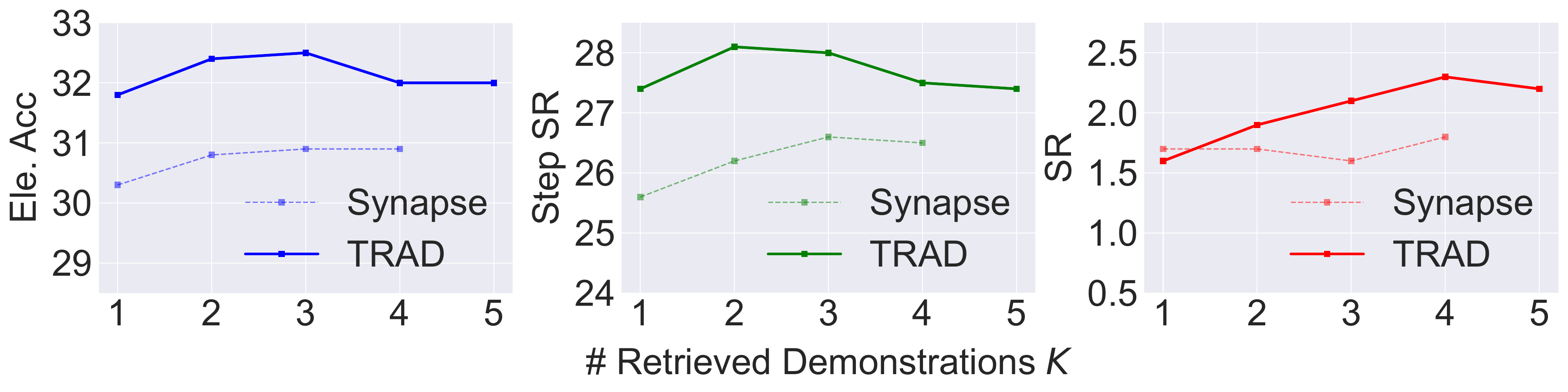}
    \vspace{-8pt}
    \caption{The effect of varying the number of retrieved demonstrations $K$ on Mind2Web benchmark. Solid lines correspond to the performance metrics of \emph{TRAD} given different $K$, and the dashed lines correspond to the \emph{Synapse} baseline. $K$ has a mild effect on the performance of \emph{TRAD} and \emph{Synapse}, and the advantage of \emph{TRAD} over \emph{Synapse} remains stable when $K$ varies.}
    \label{fig:exp-retrieval-k}
\end{figure}

Finally, we look into a common yet important hyper-parameter, the number of retrieved demonstrations $K$, and see how the advantage of \emph{TRAD} over the baseline (\emph{Synapse}) will change given different $K\in\{1,2,3,4,5\}$. We show the results over all 3 subsets in \fig{fig:exp-retrieval-k}. 
Note that the trajectory-wise prompting in \emph{Synapse} frequently exceeds the context limit when $K=5$, and thus we omit this result.

From \fig{fig:exp-retrieval-k}, we see that $K$ has a mild effect on the performance of \emph{TRAD} and \emph{Synapse}, and that the advantage of \emph{TRAD} over \emph{Synapse} consistently remains for all $K\in\{1,2,3,4\}$.

With results in \se{subsubsec:abl-temporal-extension} and \se{subsubsec:abl-k}, we now respond to our last research question.

\minisection{Answer to RQ5:} The performance and advantage of \emph{TRAD} generally remains stable with different hyper-parameter choices, i.e., temporal expansion steps, number of retrieved demonstrations. Its performance and advantage only degrade when using long backward extension, which is possibly due to the fact that historical information has already been incorporated in thoughts and does not provide further help for decision-making.

\subsection{Case Studies}\label{subsec:case-study}
\begin{figure*}[tbp]
  \centering
  \vspace{-4pt}
  \begin{subfigure}[b]{\textwidth}
    \centering
    \includegraphics[width=0.94\linewidth, height=0.44\linewidth]{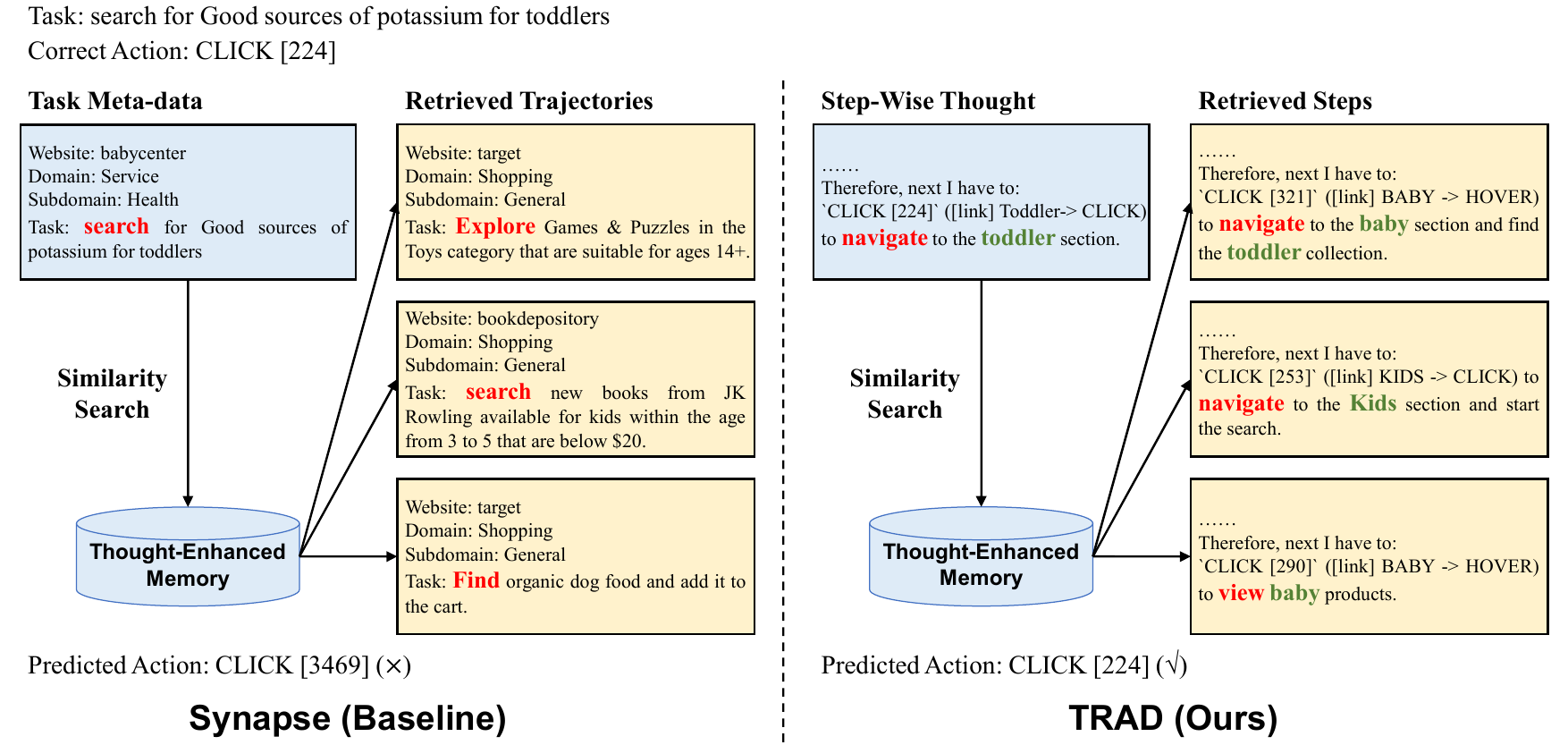}
    \caption{Representative Case 1}
    \label{fig:case-study-1}
  \end{subfigure}\\
  \begin{subfigure}[b]{\textwidth}
    \centering
    \includegraphics[width=0.94\linewidth, height=0.44\linewidth]{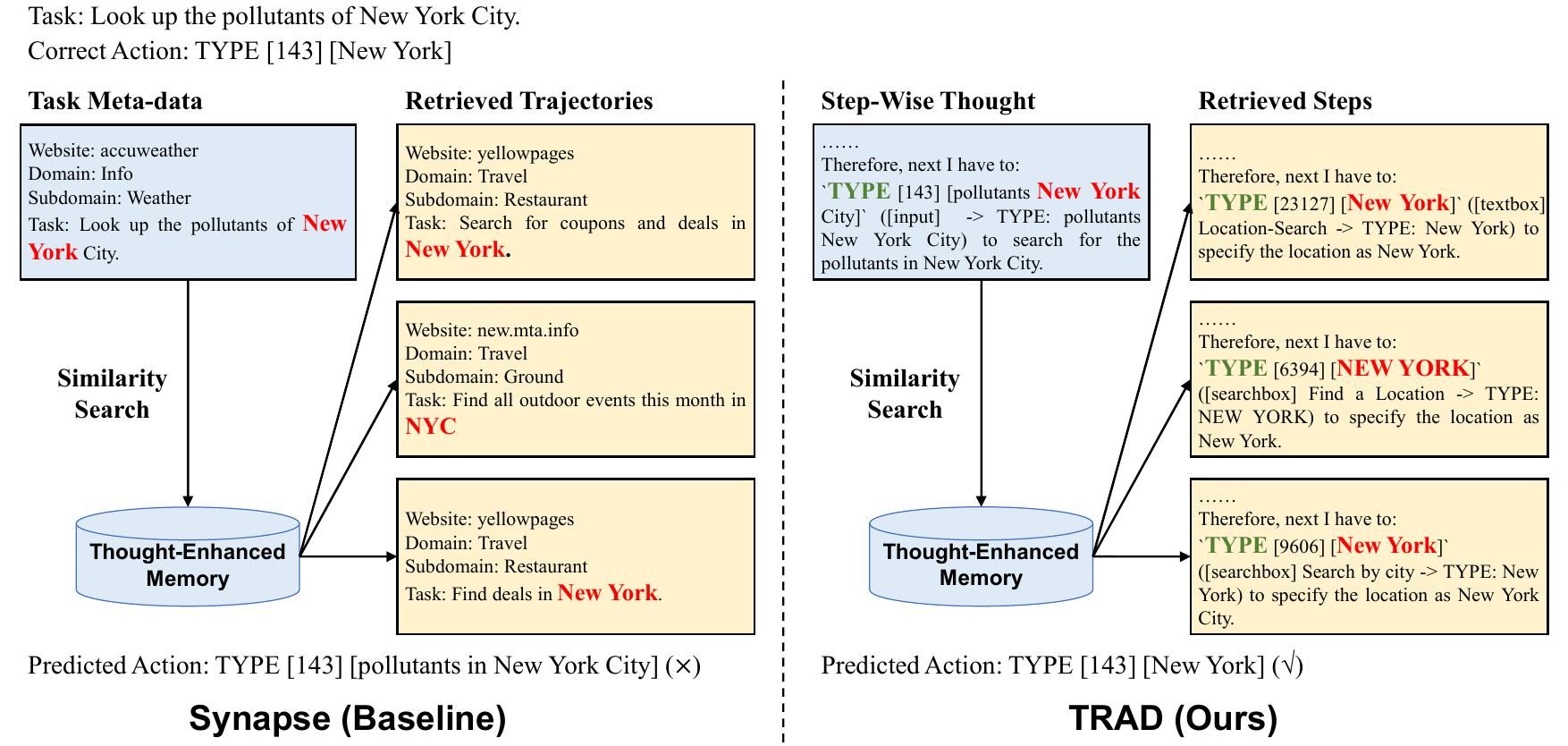}
    \caption{Representative Case 2}
    \label{fig:case-study-2}
  \end{subfigure}
  \vspace{-18pt}
  \caption{Comparison between Synapse trajectory-wise retrieval with task meta-data and TRAD step-wise retrieval with thought. (a) The trajectory-wise retrieval of \emph{Synapse} only considers \textcolor{red}{\textbf{``search''}} in task instructions and the retrieved trajectories are completely irrelevant. However, by generating thoughts with these irrelevant trajectories, \emph{thought retrieval} finds more relevant step-wise demonstrations related to \textcolor{darkgreen}{\textbf{baby (toddler)}} and \textcolor{red}{\textbf{navigation}}. (b) The trajectory-wise retrieval of \emph{Synapse} retrieves plausible examples which do not \textcolor{darkgreen}{\textbf{type}} in a text box with task meta-data. Although thoughts are imperfect, \emph{thought retrieval} finds more relevant demonstrations and \emph{TRAD} learns to input \textcolor{red}{\textbf{``New York''}}.}
  \vspace{-4pt}
  \label{fig:case-study}
\end{figure*}

At the end of this section, we present some representative trajectories or steps, where we can intuitively learn the advantages of \emph{TRAD}. We show two cases produced by \emph{Synapse} and our \emph{TRAD} agent on the cross-domain subset of Mind2Web in \fig{fig:case-study}, to demonstrate: 1) the difference between task meta-data retrieval and \emph{thought retrieval}; 2) the reason for retrieval rather than direct prediction with thought and the tolerance for imperfect thoughts.

In \fig{fig:case-study-1}, the trajectory-wise retrieval of \emph{Synapse} is obviously problematic, which only considers \textcolor{red}{\textbf{``search''}} in task instructions and the retrieved trajectories are completely irrelevant to the current one. However, when we use these irrelevant demonstrations for thought production and conduct \emph{thought retrieval} afterwards, the retrieved demonstrations become much more relevant as they all relate to \textcolor{darkgreen}{\textbf{baby (toddler)}} and reflect the process of interacting with \textcolor{red}{\textbf{navigation}} links or buttons to unfold invisible web pages during web browsing. With the demonstrations from \emph{thought retrieval}, \emph{TRAD} is capable of making the correct decision.

In \fig{fig:case-study-2}, both \emph{Synapse} and \emph{TRAD} seem to retrieve relevant examples trying to find something in \textcolor{red}{\textbf{New York}}, but if we examine the trajectories retrieved by task meta-data, 2/3 of them fulfill the condition ``New York'' by clicking some link or button rather than \textcolor{darkgreen}{\textbf{typing}} in a text box. Unfortunately, the correct action under the current state is typing, not clicking, and thus \emph{Synapse} fails to type the correct content. On the contrary, \emph{TRAD} learns to type the correct content ``New York'' into the text box, even if its thought is incorrect. This also validates our hypothesis that using thought for retrieval instead of prediction helps to correct imperfect thoughts.

\section{Real-World Deployment of TRAD}

Since Dec. 2023, we have deployed our \emph{TRAD} agent to automate some real-world office tasks in a mainstream insurance company, which owns a global business with approximately 170 million customers worldwide. We select 4 different websites and collect 100 expert trajectories for some representative tasks on each website as our memory. For evaluation, we collect 20 unseen tasks on each website, using step success rate (Step SR) and trajectory success rate (SR) as evaluation metrics. Tasks involve filling in insurance inquiry forms, implementing advanced information retrieval, etc. Since the websites are complex and contain thousands of web elements, prompting with complete trajectories is not available, hence we only consider single-step prompting with historical actions as auxiliary information.

To verify the effectiveness of \emph{TRAD}, we use two different \emph{ReAct} agents that the company has attempted as our baseline:
\begin{itemize}
    \item \emph{ReAct}-RD: randomly selects expert steps \textbf{in random trajectories} as demonstrations.
    \item \emph{ReAct}-RV: randomly selects expert steps \textbf{in relevant trajectories} retrieved by task instruction as demonstrations.
\end{itemize}
To be specific, the difference between \emph{TRAD} and \emph{ReAct}-RV is using thought for a second-time step retrieval and the aligned decision module. To further investigate the effect of \emph{thought retrieval} and \emph{aligned decision}, we also deploy a TR agent which removes our \emph{aligned decision} method, namely the \emph{TRAD} w/o TE baseline in \tb{tab:exp-mind2web-abl}. We list the results in \tb{tab:deploy}.

\begin{table}[htbp]
    \caption{Evaluation results on real-world websites from a mainstream global business insurance company.}
    \label{tab:deploy}
    \vspace{-8pt}
    \resizebox{\columnwidth}{!}{
    \begin{tabular}{cccccc}
    \toprule
    \multicolumn{2}{c}{\textbf{Method}} & ReAct-RD & ReAct-RV & TR & TRAD (Ours) \\
    \midrule
    \textbf{Website 1} & \textbf{Step SR} & 0.843 & 0.826 & 0.941 & \textbf{0.950} \\
    \textbf{(form filling)} & \textbf{SR} & 0.500 & 0.450 & \textbf{0.800} & \textbf{0.800} \\
    \midrule
    \textbf{Website 2} & \textbf{Step SR} & 0.941 & 0.937 & 0.958 & \textbf{0.974} \\
    \textbf{(advanced IR)} & \textbf{SR} & \textbf{0.900} & 0.850 & 0.850 & \textbf{0.900} \\
    \midrule
    \textbf{Website 3} & \textbf{Step SR} & 0.962 & 0.987 & \textbf{1.000} & \textbf{1.000} \\
    \textbf{(advanced IR)} & \textbf{SR} & 0.850 & 0.800 & 0.850 & \textbf{1.000} \\
    \midrule
    \textbf{Website 4} & \textbf{Step SR} & 0.820 & 0.860 & 0.845 & \textbf{1.000} \\
    \textbf{(form filling)} & \textbf{SR} & 0.350 & 0.350 & 0.400 & \textbf{1.000} \\
    \midrule
    \multirow{2}{*}{\textbf{Average}} & \textbf{Step SR} & 0.891 & 0.902 & 0.936 & \textbf{0.981} \\
    & \textbf{SR} & 0.650 & 0.613 & 0.725 & \textbf{0.925} \\
    \bottomrule
    \end{tabular}
    }
\end{table}

As can be seen in \tb{tab:deploy}, \emph{TRAD} achieves the best performance on all 4 websites, showing its advantage can remain when deployed to real-world scenarios. Moreover, we observe that \emph{TRAD} w/o TE baseline also outperforms both \emph{ReAct} agents, but exhibits noticeable disadvantages compared to the complete \emph{TRAD} agents. This justifies our design of both \emph{thought retrieval} and \emph{aligned decision}.

\minisection{Inference efficiency of \emph{TRAD}.} At inference time, our \emph{TRAD} agent only introduces little extra time consumption in \emph{thought retrieval} compared to \emph{ReAct}. We profile the inference process of \emph{TRAD} and \emph{ReAct} on all websites and tasks, and in average \emph{TRAD} takes only 11.7\% more time than \emph{ReAct}-RD, which indicates that our method achieves improvement without much sacrifice on efficiency.
\section{Discussions}

\subsection{Limitations of TRAD}
Although \emph{TRAD} exhibits excellent performances over a diverse set of tasks, it still has limitations like dependence on high-quality thought and trade-off between information and noise in \emph{temporal expansion}, and we briefly discuss about them here.

\subsubsection{Dependence on high-quality thought.}
\emph{TRAD} alleviates the issue of imperfect thoughts by its \emph{aligned decision} module, but its capability still depends heavily on the quality of thoughts and the capability of backbone LLM. To make such a step-wise retrieval-augmented method work well, the abstraction of current state is critical since it serves as the query and key for retrieval, hence the LLM used to build a \emph{TRAD} agent should at least have a decent understanding of the task.

\subsubsection{Trade-off in temporal expansion.}
\emph{TRAD} expects to keep relevant information but reduce irrelevant input context by step-wise \emph{thought retrieval}, while preserving some chance for correcting imperfect thoughts by \emph{temporal expansion}. Here exists a trade-off: a longer \emph{temporal expansion} brings not only more tolerance to imperfect thoughts, but also more irrelevant noise in demonstrations. This trade-off requires careful consideration for different tasks.

\subsection{Future Directions}
While ablation studies have been conducted to justify our design of \emph{TRAD}, there are some promising ideas worth study which can probably improve \emph{TRAD} further. We leave them as future works, and discuss them as follows.

\subsubsection{Better Demonstrations For Reasoning}
\emph{TRAD} currently employs relevant trajectories or randomly-chosen steps from them as demonstrations to generate thoughts, which still suffers from the issues discussed in \se{sec:intro} to some extent. Therefore, modifications can be made to generate thoughts of higher quality, and thus improve the overall performance of \emph{TRAD}.

\subsubsection{Better Representations For Retrieval}
As we have discussed in \se{subsec:plan-reason}, \emph{TRAD} can utilize any other methods to obtain a comprehensive abstraction of the current state in a sequential decision-making task, which can possibly serve as better queries and keys for the step-wise demonstration retrieval. Therefore, 
\emph{TRAD} can be combined with more powerful LLM planning and reasoning methods and even dense abstractions produced by LLMs pre-trained on domain-specific data like \cite{gur2023webagent}.
\section{Conclusions}

In this work, we propose a novel LLM agent augmented by step-wise demonstration retrieval (\emph{TRAD}) for sequential decision-making tasks. \emph{TRAD} first retrieves relevant step demonstrations by its thought about current state, and then complements temporally correlated steps for more informative action prediction. Extensive experiments are conducted on two different sequential decision-making tasks to validate the effectiveness of our solution, and thorough ablation studies justify the design choice and stability of our method. We further present the results from real-world deployment of our method, showing its value in real-world applications.

\begin{acks}
The Shanghai Jiao Tong University team is partially supported by Shanghai Municipal Science and Technology Major Project (2021SHZDZX0102) and National Natural Science Foundation of China (62322603, 62076161).
\end{acks}

\bibliographystyle{ACM-Reference-Format}
\bibliography{reference}


\begin{thebibliography}{48}


\ifx \showCODEN    \undefined \def \showCODEN     #1{\unskip}     \fi
\ifx \showDOI      \undefined \def \showDOI       #1{#1}\fi
\ifx \showISBNx    \undefined \def \showISBNx     #1{\unskip}     \fi
\ifx \showISBNxiii \undefined \def \showISBNxiii  #1{\unskip}     \fi
\ifx \showISSN     \undefined \def \showISSN      #1{\unskip}     \fi
\ifx \showLCCN     \undefined \def \showLCCN      #1{\unskip}     \fi
\ifx \shownote     \undefined \def \shownote      #1{#1}          \fi
\ifx \showarticletitle \undefined \def \showarticletitle #1{#1}   \fi
\ifx \showURL      \undefined \def \showURL       {\relax}        \fi
\providecommand\bibfield[2]{#2}
\providecommand\bibinfo[2]{#2}
\providecommand\natexlab[1]{#1}
\providecommand\showeprint[2][]{arXiv:#2}

\bibitem[Aeronautiques et~al\mbox{.}(1998)]%
        {aeronautiques1998pddl}
\bibfield{author}{\bibinfo{person}{Constructions Aeronautiques}, \bibinfo{person}{Adele Howe}, \bibinfo{person}{Craig Knoblock}, \bibinfo{person}{ISI~Drew McDermott}, \bibinfo{person}{Ashwin Ram}, \bibinfo{person}{Manuela Veloso}, \bibinfo{person}{Daniel Weld}, \bibinfo{person}{David~Wilkins SRI}, \bibinfo{person}{Anthony Barrett}, \bibinfo{person}{Dave Christianson}, {et~al\mbox{.}}} \bibinfo{year}{1998}\natexlab{}.
\newblock \showarticletitle{Pddl| the planning domain definition language}.
\newblock \bibinfo{journal}{\emph{Technical Report}} (\bibinfo{year}{1998}).
\newblock


\bibitem[Besta et~al\mbox{.}(2023)]%
        {besta2023got}
\bibfield{author}{\bibinfo{person}{Maciej Besta}, \bibinfo{person}{Nils Blach}, \bibinfo{person}{Ales Kubicek}, \bibinfo{person}{Robert Gerstenberger}, \bibinfo{person}{Lukas Gianinazzi}, \bibinfo{person}{Joanna Gajda}, \bibinfo{person}{Tomasz Lehmann}, \bibinfo{person}{Michal Podstawski}, \bibinfo{person}{Hubert Niewiadomski}, \bibinfo{person}{Piotr Nyczyk}, {and} \bibinfo{person}{Torsten Hoefler}.} \bibinfo{year}{2023}\natexlab{}.
\newblock \showarticletitle{Graph of Thoughts: Solving Elaborate Problems with Large Language Models}.
\newblock \bibinfo{journal}{\emph{arXiv preprint arXiv:2308.09687}} (\bibinfo{year}{2023}).
\newblock


\bibitem[Brown et~al\mbox{.}(2020)]%
        {brown2020gpt3}
\bibfield{author}{\bibinfo{person}{Tom~B. Brown}, \bibinfo{person}{Benjamin Mann}, \bibinfo{person}{Nick Ryder}, \bibinfo{person}{Melanie Subbiah}, \bibinfo{person}{Jared Kaplan}, \bibinfo{person}{Prafulla Dhariwal}, \bibinfo{person}{Arvind Neelakantan}, \bibinfo{person}{Pranav Shyam}, \bibinfo{person}{Girish Sastry}, \bibinfo{person}{Amanda Askell}, \bibinfo{person}{Sandhini Agarwal}, \bibinfo{person}{Ariel Herbert{-}Voss}, \bibinfo{person}{Gretchen Krueger}, \bibinfo{person}{Tom Henighan}, \bibinfo{person}{Rewon Child}, \bibinfo{person}{Aditya Ramesh}, \bibinfo{person}{Daniel~M. Ziegler}, \bibinfo{person}{Jeffrey Wu}, \bibinfo{person}{Clemens Winter}, \bibinfo{person}{Christopher Hesse}, \bibinfo{person}{Mark Chen}, \bibinfo{person}{Eric Sigler}, \bibinfo{person}{Mateusz Litwin}, \bibinfo{person}{Scott Gray}, \bibinfo{person}{Benjamin Chess}, \bibinfo{person}{Jack Clark}, \bibinfo{person}{Christopher Berner}, \bibinfo{person}{Sam McCandlish}, \bibinfo{person}{Alec Radford}, \bibinfo{person}{Ilya Sutskever},
  {and} \bibinfo{person}{Dario Amodei}.} \bibinfo{year}{2020}\natexlab{}.
\newblock \showarticletitle{Language Models are Few-Shot Learners}. In \bibinfo{booktitle}{\emph{Proceedings of the 34th Advances in Neural Information Processing Systems (NeurIPS)}}.
\newblock


\bibitem[Deng et~al\mbox{.}(2023)]%
        {deng2023mind2web}
\bibfield{author}{\bibinfo{person}{Xiang Deng}, \bibinfo{person}{Yu Gu}, \bibinfo{person}{Boyuan Zheng}, \bibinfo{person}{Shijie Chen}, \bibinfo{person}{Samuel Stevens}, \bibinfo{person}{Boshi Wang}, \bibinfo{person}{Huan Sun}, {and} \bibinfo{person}{Yu Su}.} \bibinfo{year}{2023}\natexlab{}.
\newblock \showarticletitle{Mind2Web: Towards a Generalist Agent for the Web}. In \bibinfo{booktitle}{\emph{Proceedings of the 37th Advances in Neural Information Processing Systems (NeurIPS)}}.
\newblock


\bibitem[Devlin et~al\mbox{.}(2018)]%
        {devlin2018bert}
\bibfield{author}{\bibinfo{person}{Jacob Devlin}, \bibinfo{person}{Ming-Wei Chang}, \bibinfo{person}{Kenton Lee}, {and} \bibinfo{person}{Kristina Toutanova}.} \bibinfo{year}{2018}\natexlab{}.
\newblock \showarticletitle{Bert: Pre-training of deep bidirectional transformers for language understanding}.
\newblock \bibinfo{journal}{\emph{arXiv preprint arXiv:1810.04805}} (\bibinfo{year}{2018}).
\newblock


\bibitem[Ding et~al\mbox{.}(2023)]%
        {ding2023xot}
\bibfield{author}{\bibinfo{person}{Ruomeng Ding}, \bibinfo{person}{Chaoyun Zhang}, \bibinfo{person}{Lu Wang}, \bibinfo{person}{Yong Xu}, \bibinfo{person}{Minghua Ma}, \bibinfo{person}{Wei Zhang}, \bibinfo{person}{Si Qin}, \bibinfo{person}{Saravan Rajmohan}, \bibinfo{person}{Qingwei Lin}, {and} \bibinfo{person}{Dongmei Zhang}.} \bibinfo{year}{2023}\natexlab{}.
\newblock \showarticletitle{Everything of thoughts: Defying the law of penrose triangle for thought generation}.
\newblock \bibinfo{journal}{\emph{arXiv preprint arXiv:2311.04254}} (\bibinfo{year}{2023}).
\newblock


\bibitem[Ferraretto et~al\mbox{.}(2023)]%
        {ferraretto2023exaranker}
\bibfield{author}{\bibinfo{person}{Fernando Ferraretto}, \bibinfo{person}{Thiago Laitz}, \bibinfo{person}{Roberto Lotufo}, {and} \bibinfo{person}{Rodrigo Nogueira}.} \bibinfo{year}{2023}\natexlab{}.
\newblock \showarticletitle{ExaRanker: Synthetic Explanations Improve Neural Rankers}. In \bibinfo{booktitle}{\emph{Proceedings of the 46th International {ACM} {SIGIR} Conference on Research and Development in Information Retrieval ({SIGIR})}}. \bibinfo{pages}{2409–--2414}.
\newblock


\bibitem[Gur et~al\mbox{.}(2024)]%
        {gur2023webagent}
\bibfield{author}{\bibinfo{person}{Izzeddin Gur}, \bibinfo{person}{Hiroki Furuta}, \bibinfo{person}{Austin Huang}, \bibinfo{person}{Mustafa Safdari}, \bibinfo{person}{Yutaka Matsuo}, \bibinfo{person}{Douglas Eck}, {and} \bibinfo{person}{Aleksandra Faust}.} \bibinfo{year}{2024}\natexlab{}.
\newblock \showarticletitle{A Real-World WebAgent with Planning, Long Context Understanding, and Program Synthesis}. In \bibinfo{booktitle}{\emph{Proceedings of The 12th International Conference on Learning Representations ({ICLR})}}.
\newblock


\bibitem[Gur et~al\mbox{.}(2023)]%
        {gur2023html}
\bibfield{author}{\bibinfo{person}{Izzeddin Gur}, \bibinfo{person}{Ofir Nachum}, \bibinfo{person}{Yingjie Miao}, \bibinfo{person}{Mustafa Safdari}, \bibinfo{person}{Austin Huang}, \bibinfo{person}{Aakanksha Chowdhery}, \bibinfo{person}{Sharan Narang}, \bibinfo{person}{Noah Fiedel}, {and} \bibinfo{person}{Aleksandra Faust}.} \bibinfo{year}{2023}\natexlab{}.
\newblock \showarticletitle{Understanding {HTML} with Large Language Models}. In \bibinfo{booktitle}{\emph{Findings of the Association for Computational Linguistics ({EMNLP})}}. \bibinfo{pages}{2803--2821}.
\newblock


\bibitem[Hao et~al\mbox{.}(2023)]%
        {hao2023rap}
\bibfield{author}{\bibinfo{person}{Shibo Hao}, \bibinfo{person}{Yi Gu}, \bibinfo{person}{Haodi Ma}, \bibinfo{person}{Joshua~Jiahua Hong}, \bibinfo{person}{Zhen Wang}, \bibinfo{person}{Daisy~Zhe Wang}, {and} \bibinfo{person}{Zhiting Hu}.} \bibinfo{year}{2023}\natexlab{}.
\newblock \showarticletitle{Reasoning with Language Model is Planning with World Model}. In \bibinfo{booktitle}{\emph{Proceedings of the 2023 Conference on Empirical Methods in Natural Language Processing ({EMNLP})}}. \bibinfo{pages}{8154--8173}.
\newblock


\bibitem[Holtzman et~al\mbox{.}(2020)]%
        {holtzman2020curious}
\bibfield{author}{\bibinfo{person}{Ari Holtzman}, \bibinfo{person}{Jan Buys}, \bibinfo{person}{Li Du}, \bibinfo{person}{Maxwell Forbes}, {and} \bibinfo{person}{Yejin Choi}.} \bibinfo{year}{2020}\natexlab{}.
\newblock \showarticletitle{The Curious Case of Neural Text Degeneration}. In \bibinfo{booktitle}{\emph{Proceedings of the 8th International Conference on Learning Representations ({ICLR})}}.
\newblock


\bibitem[Karpukhin et~al\mbox{.}(2020)]%
        {karpukhin2020dpr}
\bibfield{author}{\bibinfo{person}{Vladimir Karpukhin}, \bibinfo{person}{Barlas Oguz}, \bibinfo{person}{Sewon Min}, \bibinfo{person}{Patrick S.~H. Lewis}, \bibinfo{person}{Ledell Wu}, \bibinfo{person}{Sergey Edunov}, \bibinfo{person}{Danqi Chen}, {and} \bibinfo{person}{Wen{-}tau Yih}.} \bibinfo{year}{2020}\natexlab{}.
\newblock \showarticletitle{Dense Passage Retrieval for Open-Domain Question Answering}. In \bibinfo{booktitle}{\emph{Proceedings of the 2020 Conference on Empirical Methods in Natural Language Processing ({EMNLP})}}. \bibinfo{pages}{6769--6781}.
\newblock


\bibitem[Kim et~al\mbox{.}(2023)]%
        {kim2023rci}
\bibfield{author}{\bibinfo{person}{Geunwoo Kim}, \bibinfo{person}{Pierre Baldi}, {and} \bibinfo{person}{Stephen McAleer}.} \bibinfo{year}{2023}\natexlab{}.
\newblock \showarticletitle{Language Models can Solve Computer Tasks}. In \bibinfo{booktitle}{\emph{Proceedings of the 37th Advances in Neural Information Processing Systems (NeurIPS)}}.
\newblock


\bibitem[Liang et~al\mbox{.}(2023)]%
        {liang2023code}
\bibfield{author}{\bibinfo{person}{Jacky Liang}, \bibinfo{person}{Wenlong Huang}, \bibinfo{person}{Fei Xia}, \bibinfo{person}{Peng Xu}, \bibinfo{person}{Karol Hausman}, \bibinfo{person}{Brian Ichter}, \bibinfo{person}{Pete Florence}, {and} \bibinfo{person}{Andy Zeng}.} \bibinfo{year}{2023}\natexlab{}.
\newblock \showarticletitle{Code as Policies: Language Model Programs for Embodied Control}. In \bibinfo{booktitle}{\emph{Proceedings of 2023 IEEE International Conference on Robotics and Automation ({ICRA})}}. \bibinfo{pages}{9493--9500}.
\newblock


\bibitem[Liu et~al\mbox{.}(2023)]%
        {liu2023llm+p}
\bibfield{author}{\bibinfo{person}{Bo Liu}, \bibinfo{person}{Yuqian Jiang}, \bibinfo{person}{Xiaohan Zhang}, \bibinfo{person}{Qiang Liu}, \bibinfo{person}{Shiqi Zhang}, \bibinfo{person}{Joydeep Biswas}, {and} \bibinfo{person}{Peter Stone}.} \bibinfo{year}{2023}\natexlab{}.
\newblock \showarticletitle{LLM+P: Empowering large language models with optimal planning proficiency}.
\newblock \bibinfo{journal}{\emph{arXiv preprint arXiv:2304.11477}} (\bibinfo{year}{2023}).
\newblock


\bibitem[Liu et~al\mbox{.}(2021)]%
        {liu2021good-example}
\bibfield{author}{\bibinfo{person}{Jiachang Liu}, \bibinfo{person}{Dinghan Shen}, \bibinfo{person}{Yizhe Zhang}, \bibinfo{person}{Bill Dolan}, \bibinfo{person}{Lawrence Carin}, {and} \bibinfo{person}{Weizhu Chen}.} \bibinfo{year}{2021}\natexlab{}.
\newblock \showarticletitle{What Makes Good In-Context Examples for GPT-3?}
\newblock \bibinfo{journal}{\emph{arXiv preprint arXiv:2101.06804}} (\bibinfo{year}{2021}).
\newblock


\bibitem[Mackie et~al\mbox{.}(2023)]%
        {mackie2023grf}
\bibfield{author}{\bibinfo{person}{Iain Mackie}, \bibinfo{person}{Shubham Chatterjee}, {and} \bibinfo{person}{Jeffrey Dalton}.} \bibinfo{year}{2023}\natexlab{}.
\newblock \showarticletitle{Generative Relevance Feedback with Large Language Models}. In \bibinfo{booktitle}{\emph{Proceedings of the 46th International {ACM} {SIGIR} Conference on Research and Development in Information Retrieval ({SIGIR})}}. \bibinfo{pages}{2026--2031}.
\newblock


\bibitem[Nakano et~al\mbox{.}(2021)]%
        {nakano2021webgpt}
\bibfield{author}{\bibinfo{person}{Reiichiro Nakano}, \bibinfo{person}{Jacob Hilton}, \bibinfo{person}{Suchir Balaji}, \bibinfo{person}{Jeff Wu}, \bibinfo{person}{Long Ouyang}, \bibinfo{person}{Christina Kim}, \bibinfo{person}{Christopher Hesse}, \bibinfo{person}{Shantanu Jain}, \bibinfo{person}{Vineet Kosaraju}, \bibinfo{person}{William Saunders}, {et~al\mbox{.}}} \bibinfo{year}{2021}\natexlab{}.
\newblock \showarticletitle{Webgpt: Browser-assisted question-answering with human feedback}.
\newblock \bibinfo{journal}{\emph{arXiv preprint arXiv:2112.09332}} (\bibinfo{year}{2021}).
\newblock


\bibitem[OpenAI(2023)]%
        {openai2023gpt4}
\bibfield{author}{\bibinfo{person}{OpenAI}.} \bibinfo{year}{2023}\natexlab{}.
\newblock \showarticletitle{GPT-4 Technical Report}.
\newblock \bibinfo{journal}{\emph{arXiv preprint arXiv:2303.08774}} (\bibinfo{year}{2023}).
\newblock


\bibitem[Ouyang et~al\mbox{.}(2022)]%
        {ouyang2022training}
\bibfield{author}{\bibinfo{person}{Long Ouyang}, \bibinfo{person}{Jeffrey Wu}, \bibinfo{person}{Xu Jiang}, \bibinfo{person}{Diogo Almeida}, \bibinfo{person}{Carroll Wainwright}, \bibinfo{person}{Pamela Mishkin}, \bibinfo{person}{Chong Zhang}, \bibinfo{person}{Sandhini Agarwal}, \bibinfo{person}{Katarina Slama}, \bibinfo{person}{Alex Ray}, {et~al\mbox{.}}} \bibinfo{year}{2022}\natexlab{}.
\newblock \showarticletitle{Training language models to follow instructions with human feedback}. In \bibinfo{booktitle}{\emph{Proceedings of the 36th Advances in Neural Information Processing Systems (NeurIPS)}}. \bibinfo{pages}{27730--27744}.
\newblock


\bibitem[Park et~al\mbox{.}(2023)]%
        {park2023generative}
\bibfield{author}{\bibinfo{person}{Joon~Sung Park}, \bibinfo{person}{Joseph O'Brien}, \bibinfo{person}{Carrie~Jun Cai}, \bibinfo{person}{Meredith~Ringel Morris}, \bibinfo{person}{Percy Liang}, {and} \bibinfo{person}{Michael~S Bernstein}.} \bibinfo{year}{2023}\natexlab{}.
\newblock \showarticletitle{Generative agents: Interactive simulacra of human behavior}. In \bibinfo{booktitle}{\emph{Proceedings of the 36th Annual ACM Symposium on User Interface Software and Technology ({UIST})}}. \bibinfo{pages}{1--22}.
\newblock


\bibitem[Radford et~al\mbox{.}(2019)]%
        {radford2019gpt2}
\bibfield{author}{\bibinfo{person}{Alec Radford}, \bibinfo{person}{Jeffrey Wu}, \bibinfo{person}{Rewon Child}, \bibinfo{person}{David Luan}, \bibinfo{person}{Dario Amodei}, {and} \bibinfo{person}{Ilya Sutskever}.} \bibinfo{year}{2019}\natexlab{}.
\newblock \showarticletitle{Language models are unsupervised multitask learners}.
\newblock \bibinfo{journal}{\emph{OpenAI Blog}} (\bibinfo{year}{2019}).
\newblock


\bibitem[Reimers and Gurevych(2019)]%
        {reimers2019sbert}
\bibfield{author}{\bibinfo{person}{Nils Reimers} {and} \bibinfo{person}{Iryna Gurevych}.} \bibinfo{year}{2019}\natexlab{}.
\newblock \showarticletitle{Sentence-BERT: Sentence Embeddings using Siamese BERT-Networks}. In \bibinfo{booktitle}{\emph{Proceedings of the 2019 Conference on Empirical Methods in Natural Language Processing and the 9th International Joint Conference on Natural Language Processing ({EMNLP-IJCNLP})}}. \bibinfo{pages}{3980--3990}.
\newblock


\bibitem[Roziere et~al\mbox{.}(2023)]%
        {roziere2023codellama}
\bibfield{author}{\bibinfo{person}{Baptiste Roziere}, \bibinfo{person}{Jonas Gehring}, \bibinfo{person}{Fabian Gloeckle}, \bibinfo{person}{Sten Sootla}, \bibinfo{person}{Itai Gat}, \bibinfo{person}{Xiaoqing~Ellen Tan}, \bibinfo{person}{Yossi Adi}, \bibinfo{person}{Jingyu Liu}, \bibinfo{person}{Tal Remez}, \bibinfo{person}{J{\'e}r{\'e}my Rapin}, {et~al\mbox{.}}} \bibinfo{year}{2023}\natexlab{}.
\newblock \showarticletitle{Code llama: Open foundation models for code}.
\newblock \bibinfo{journal}{\emph{arXiv preprint arXiv:2308.12950}} (\bibinfo{year}{2023}).
\newblock


\bibitem[Rubin et~al\mbox{.}(2022)]%
        {rubin2022learning}
\bibfield{author}{\bibinfo{person}{Ohad Rubin}, \bibinfo{person}{Jonathan Herzig}, {and} \bibinfo{person}{Jonathan Berant}.} \bibinfo{year}{2022}\natexlab{}.
\newblock \showarticletitle{Learning To Retrieve Prompts for In-Context Learning}. In \bibinfo{booktitle}{\emph{Proceedings of the 2022 Conference of the North American Chapter of the Association for Computational Linguistics: Human Language Technologies ({NAACL-HLT})}}. \bibinfo{pages}{2655--2671}.
\newblock


\bibitem[Schick et~al\mbox{.}(2023)]%
        {schick2023toolformer}
\bibfield{author}{\bibinfo{person}{Timo Schick}, \bibinfo{person}{Jane Dwivedi-Yu}, \bibinfo{person}{Roberto Dess{\`\i}}, \bibinfo{person}{Roberta Raileanu}, \bibinfo{person}{Maria Lomeli}, \bibinfo{person}{Luke Zettlemoyer}, \bibinfo{person}{Nicola Cancedda}, {and} \bibinfo{person}{Thomas Scialom}.} \bibinfo{year}{2023}\natexlab{}.
\newblock \showarticletitle{Toolformer: Language models can teach themselves to use tools}. In \bibinfo{booktitle}{\emph{Proceedings of the 37th Advances in Neural Information Processing Systems (NeurIPS)}}.
\newblock


\bibitem[Shi et~al\mbox{.}(2017)]%
        {shi2017miniwob}
\bibfield{author}{\bibinfo{person}{Tianlin Shi}, \bibinfo{person}{Andrej Karpathy}, \bibinfo{person}{Linxi Fan}, \bibinfo{person}{Jonathan Hernandez}, {and} \bibinfo{person}{Percy Liang}.} \bibinfo{year}{2017}\natexlab{}.
\newblock \showarticletitle{World of Bits: An Open-Domain Platform for Web-Based Agents}. In \bibinfo{booktitle}{\emph{Proceedings of the 34th International Conference on Machine Learning ({ICML})}}, Vol.~\bibinfo{volume}{70}. \bibinfo{pages}{3135--3144}.
\newblock


\bibitem[Shinn et~al\mbox{.}(2023)]%
        {shinn2023reflexion}
\bibfield{author}{\bibinfo{person}{Noah Shinn}, \bibinfo{person}{Federico Cassano}, \bibinfo{person}{Ashwin Gopinath}, \bibinfo{person}{Karthik~R Narasimhan}, {and} \bibinfo{person}{Shunyu Yao}.} \bibinfo{year}{2023}\natexlab{}.
\newblock \showarticletitle{Reflexion: Language agents with verbal reinforcement learning}. In \bibinfo{booktitle}{\emph{Proceedings of the 37th Advances in Neural Information Processing Systems (NeurIPS)}}.
\newblock


\bibitem[Shridhar et~al\mbox{.}(2020)]%
        {shridhar2020alfred}
\bibfield{author}{\bibinfo{person}{Mohit Shridhar}, \bibinfo{person}{Jesse Thomason}, \bibinfo{person}{Daniel Gordon}, \bibinfo{person}{Yonatan Bisk}, \bibinfo{person}{Winson Han}, \bibinfo{person}{Roozbeh Mottaghi}, \bibinfo{person}{Luke Zettlemoyer}, {and} \bibinfo{person}{Dieter Fox}.} \bibinfo{year}{2020}\natexlab{}.
\newblock \showarticletitle{{ALFRED:} {A} Benchmark for Interpreting Grounded Instructions for Everyday Tasks}. In \bibinfo{booktitle}{\emph{Proceedings of the 2020 {IEEE/CVF} Conference on Computer Vision and Pattern Recognition ({CVPR})}}. \bibinfo{pages}{10737--10746}.
\newblock


\bibitem[Shridhar et~al\mbox{.}(2021)]%
        {shridhar2021alfworld}
\bibfield{author}{\bibinfo{person}{Mohit Shridhar}, \bibinfo{person}{Xingdi Yuan}, \bibinfo{person}{Marc{-}Alexandre C{\^{o}}t{\'{e}}}, \bibinfo{person}{Yonatan Bisk}, \bibinfo{person}{Adam Trischler}, {and} \bibinfo{person}{Matthew~J. Hausknecht}.} \bibinfo{year}{2021}\natexlab{}.
\newblock \showarticletitle{ALFWorld: Aligning Text and Embodied Environments for Interactive Learning}. In \bibinfo{booktitle}{\emph{Proceedings of 9th International Conference on Learning Representations ({ICLR})}}.
\newblock


\bibitem[Team(2023)]%
        {lmsys2023longchat}
\bibfield{author}{\bibinfo{person}{The~LongChat Team}.} \bibinfo{year}{2023}\natexlab{}.
\newblock \bibinfo{booktitle}{\emph{How Long Can Open-Source LLMs Truly Promise on Context Length?}}
\newblock
\urldef\tempurl%
\url{https://lmsys.org/blog/2023-06-29-longchat/}
\showURL{%
\tempurl}


\bibitem[Touvron et~al\mbox{.}(2023)]%
        {touvron2023llama}
\bibfield{author}{\bibinfo{person}{Hugo Touvron}, \bibinfo{person}{Thibaut Lavril}, \bibinfo{person}{Gautier Izacard}, \bibinfo{person}{Xavier Martinet}, \bibinfo{person}{Marie-Anne Lachaux}, \bibinfo{person}{Timothée Lacroix}, \bibinfo{person}{Baptiste Rozière}, \bibinfo{person}{Naman Goyal}, \bibinfo{person}{Eric Hambro}, \bibinfo{person}{Faisal Azhar}, \bibinfo{person}{Aurelien Rodriguez}, \bibinfo{person}{Armand Joulin}, \bibinfo{person}{Edouard Grave}, {and} \bibinfo{person}{Guillaume Lample}.} \bibinfo{year}{2023}\natexlab{}.
\newblock \showarticletitle{LLaMA: Open and Efficient Foundation Language Models}.
\newblock \bibinfo{journal}{\emph{arXiv preprint arXiv:2302.13971}} (\bibinfo{year}{2023}).
\newblock


\bibitem[Trivedi et~al\mbox{.}(2023)]%
        {trivedi2023ircot}
\bibfield{author}{\bibinfo{person}{Harsh Trivedi}, \bibinfo{person}{Niranjan Balasubramanian}, \bibinfo{person}{Tushar Khot}, {and} \bibinfo{person}{Ashish Sabharwal}.} \bibinfo{year}{2023}\natexlab{}.
\newblock \showarticletitle{Interleaving Retrieval with Chain-of-Thought Reasoning for Knowledge-Intensive Multi-Step Questions}. In \bibinfo{booktitle}{\emph{Proceedings of the 61st Annual Meeting of the Association for Computational Linguistics ({ACL})}}. \bibinfo{pages}{10014--10037}.
\newblock


\bibitem[Wang et~al\mbox{.}(2023d)]%
        {wang2023voyager}
\bibfield{author}{\bibinfo{person}{Guanzhi Wang}, \bibinfo{person}{Yuqi Xie}, \bibinfo{person}{Yunfan Jiang}, \bibinfo{person}{Ajay Mandlekar}, \bibinfo{person}{Chaowei Xiao}, \bibinfo{person}{Yuke Zhu}, \bibinfo{person}{Linxi Fan}, {and} \bibinfo{person}{Anima Anandkumar}.} \bibinfo{year}{2023}\natexlab{d}.
\newblock \showarticletitle{Voyager: An open-ended embodied agent with large language models}.
\newblock \bibinfo{journal}{\emph{arXiv preprint arXiv:2305.16291}} (\bibinfo{year}{2023}).
\newblock


\bibitem[Wang et~al\mbox{.}(2023b)]%
        {wang2023survey}
\bibfield{author}{\bibinfo{person}{Lei Wang}, \bibinfo{person}{Chen Ma}, \bibinfo{person}{Xueyang Feng}, \bibinfo{person}{Zeyu Zhang}, \bibinfo{person}{Hao Yang}, \bibinfo{person}{Jingsen Zhang}, \bibinfo{person}{Zhiyuan Chen}, \bibinfo{person}{Jiakai Tang}, \bibinfo{person}{Xu Chen}, \bibinfo{person}{Yankai Lin}, {et~al\mbox{.}}} \bibinfo{year}{2023}\natexlab{b}.
\newblock \showarticletitle{A survey on large language model based autonomous agents}.
\newblock \bibinfo{journal}{\emph{arXiv preprint arXiv:2308.11432}} (\bibinfo{year}{2023}).
\newblock


\bibitem[Wang et~al\mbox{.}(2023c)]%
        {wang2023sc}
\bibfield{author}{\bibinfo{person}{Xuezhi Wang}, \bibinfo{person}{Jason Wei}, \bibinfo{person}{Dale Schuurmans}, \bibinfo{person}{Quoc~V. Le}, \bibinfo{person}{Ed~H. Chi}, \bibinfo{person}{Sharan Narang}, \bibinfo{person}{Aakanksha Chowdhery}, {and} \bibinfo{person}{Denny Zhou}.} \bibinfo{year}{2023}\natexlab{c}.
\newblock \showarticletitle{Self-Consistency Improves Chain of Thought Reasoning in Language Models}. In \bibinfo{booktitle}{\emph{The 11th International Conference on Learning Representations, ({ICLR})}}.
\newblock


\bibitem[Wang et~al\mbox{.}(2023a)]%
        {wang2023deps}
\bibfield{author}{\bibinfo{person}{Zihao Wang}, \bibinfo{person}{Shaofei Cai}, \bibinfo{person}{Anji Liu}, \bibinfo{person}{Xiaojian Ma}, {and} \bibinfo{person}{Yitao Liang}.} \bibinfo{year}{2023}\natexlab{a}.
\newblock \showarticletitle{Describe, explain, plan and select: Interactive planning with large language models enables open-world multi-task agents}. In \bibinfo{booktitle}{\emph{Proceedings of the 37th Advances in Neural Information Processing Systems (NeurIPS)}}.
\newblock


\bibitem[Wei et~al\mbox{.}(2022)]%
        {wei2022cot}
\bibfield{author}{\bibinfo{person}{Jason Wei}, \bibinfo{person}{Xuezhi Wang}, \bibinfo{person}{Dale Schuurmans}, \bibinfo{person}{Maarten Bosma}, \bibinfo{person}{Brian Ichter}, \bibinfo{person}{Fei Xia}, \bibinfo{person}{Ed~H. Chi}, \bibinfo{person}{Quoc~V. Le}, {and} \bibinfo{person}{Denny Zhou}.} \bibinfo{year}{2022}\natexlab{}.
\newblock \showarticletitle{Chain-of-Thought Prompting Elicits Reasoning in Large Language Models}. In \bibinfo{booktitle}{\emph{Proceedings of the 36th Advances in Neural Information Processing Systems (NeurIPS)}}.
\newblock


\bibitem[Wu et~al\mbox{.}(2023)]%
        {wu2023adaptive}
\bibfield{author}{\bibinfo{person}{Zhiyong Wu}, \bibinfo{person}{Yaoxiang Wang}, \bibinfo{person}{Jiacheng Ye}, {and} \bibinfo{person}{Lingpeng Kong}.} \bibinfo{year}{2023}\natexlab{}.
\newblock \showarticletitle{Self-Adaptive In-Context Learning: An Information Compression Perspective for In-Context Example Selection and Ordering}. In \bibinfo{booktitle}{\emph{Proceedings of the 61st Annual Meeting of the Association for Computational Linguistics ({ACL})}}. \bibinfo{pages}{1423--1436}.
\newblock


\bibitem[Yao et~al\mbox{.}(2022)]%
        {yao2022webshop}
\bibfield{author}{\bibinfo{person}{Shunyu Yao}, \bibinfo{person}{Howard Chen}, \bibinfo{person}{John Yang}, {and} \bibinfo{person}{Karthik Narasimhan}.} \bibinfo{year}{2022}\natexlab{}.
\newblock \showarticletitle{WebShop: Towards Scalable Real-World Web Interaction with Grounded Language Agents}. In \bibinfo{booktitle}{\emph{Proceedings of 36th Conference on Neural Information Processing Systems (NeurIPS)}}.
\newblock


\bibitem[Yao et~al\mbox{.}(2023a)]%
        {yao2023tot}
\bibfield{author}{\bibinfo{person}{Shunyu Yao}, \bibinfo{person}{Dian Yu}, \bibinfo{person}{Jeffrey Zhao}, \bibinfo{person}{Izhak Shafran}, \bibinfo{person}{Thomas~L. Griffiths}, \bibinfo{person}{Yuan Cao}, {and} \bibinfo{person}{Karthik Narasimhan}.} \bibinfo{year}{2023}\natexlab{a}.
\newblock \showarticletitle{Tree of Thoughts: Deliberate Problem Solving with Large Language Models}. In \bibinfo{booktitle}{\emph{Proceedings of 37th Conference on Neural Information Processing Systems (NeurIPS)}}.
\newblock


\bibitem[Yao et~al\mbox{.}(2023b)]%
        {yao2023react}
\bibfield{author}{\bibinfo{person}{Shunyu Yao}, \bibinfo{person}{Jeffrey Zhao}, \bibinfo{person}{Dian Yu}, \bibinfo{person}{Nan Du}, \bibinfo{person}{Izhak Shafran}, \bibinfo{person}{Karthik~R. Narasimhan}, {and} \bibinfo{person}{Yuan Cao}.} \bibinfo{year}{2023}\natexlab{b}.
\newblock \showarticletitle{ReAct: Synergizing Reasoning and Acting in Language Models}. In \bibinfo{booktitle}{\emph{Proceedings of The 11th International Conference on Learning Representations ({ICLR})}}.
\newblock


\bibitem[Ye et~al\mbox{.}(2023)]%
        {ye2023dater}
\bibfield{author}{\bibinfo{person}{Yunhu Ye}, \bibinfo{person}{Binyuan Hui}, \bibinfo{person}{Min Yang}, \bibinfo{person}{Binhua Li}, \bibinfo{person}{Fei Huang}, {and} \bibinfo{person}{Yongbin Li}.} \bibinfo{year}{2023}\natexlab{}.
\newblock \showarticletitle{Large Language Models are Versatile Decomposers: Decomposing Evidence and Questions for Table-based Reasoning}. In \bibinfo{booktitle}{\emph{Proceedings of the 46th International {ACM} {SIGIR} Conference on Research and Development in Information Retrieval ({SIGIR})}}. \bibinfo{pages}{174--184}.
\newblock


\bibitem[Zhang et~al\mbox{.}(2022)]%
        {zhang2022unlabeled}
\bibfield{author}{\bibinfo{person}{Yiming Zhang}, \bibinfo{person}{Shi Feng}, {and} \bibinfo{person}{Chenhao Tan}.} \bibinfo{year}{2022}\natexlab{}.
\newblock \showarticletitle{Active Example Selection for In-Context Learning}. In \bibinfo{booktitle}{\emph{Proceedings of the 2022 Conference on Empirical Methods in Natural Language Processing ({EMNLP})}}. \bibinfo{pages}{9134--9148}.
\newblock


\bibitem[Zheng et~al\mbox{.}(2024a)]%
        {zheng2023stepback}
\bibfield{author}{\bibinfo{person}{Huaixiu~Steven Zheng}, \bibinfo{person}{Swaroop Mishra}, \bibinfo{person}{Xinyun Chen}, \bibinfo{person}{Heng-Tze Cheng}, \bibinfo{person}{Ed~H. Chi}, \bibinfo{person}{Quoc~V Le}, {and} \bibinfo{person}{Denny Zhou}.} \bibinfo{year}{2024}\natexlab{a}.
\newblock \showarticletitle{Step-Back Prompting Enables Reasoning Via Abstraction in Large Language Models}. In \bibinfo{booktitle}{\emph{Proceedings of The 12th International Conference on Learning Representations ({ICLR})}}.
\newblock


\bibitem[Zheng et~al\mbox{.}(2024b)]%
        {zheng2023synapse}
\bibfield{author}{\bibinfo{person}{Longtao Zheng}, \bibinfo{person}{Rundong Wang}, \bibinfo{person}{Xinrun Wang}, {and} \bibinfo{person}{Bo An}.} \bibinfo{year}{2024}\natexlab{b}.
\newblock \showarticletitle{Synapse: Trajectory-as-Exemplar Prompting with Memory for Computer Control}. In \bibinfo{booktitle}{\emph{Proceedings of 12th International Conference on Learning Representations ({ICLR})}}.
\newblock


\bibitem[Zhou et~al\mbox{.}(2023)]%
        {zhou2023least}
\bibfield{author}{\bibinfo{person}{Denny Zhou}, \bibinfo{person}{Nathanael Sch{\"{a}}rli}, \bibinfo{person}{Le Hou}, \bibinfo{person}{Jason Wei}, \bibinfo{person}{Nathan Scales}, \bibinfo{person}{Xuezhi Wang}, \bibinfo{person}{Dale Schuurmans}, \bibinfo{person}{Claire Cui}, \bibinfo{person}{Olivier Bousquet}, \bibinfo{person}{Quoc~V. Le}, {and} \bibinfo{person}{Ed~H. Chi}.} \bibinfo{year}{2023}\natexlab{}.
\newblock \showarticletitle{Least-to-Most Prompting Enables Complex Reasoning in Large Language Models}. In \bibinfo{booktitle}{\emph{The 11th International Conference on Learning Representations ({ICLR})}}.
\newblock


\bibitem[Zhu et~al\mbox{.}(2023)]%
        {zhu2023large}
\bibfield{author}{\bibinfo{person}{Yutao Zhu}, \bibinfo{person}{Huaying Yuan}, \bibinfo{person}{Shuting Wang}, \bibinfo{person}{Jiongnan Liu}, \bibinfo{person}{Wenhan Liu}, \bibinfo{person}{Chenlong Deng}, \bibinfo{person}{Zhicheng Dou}, {and} \bibinfo{person}{Ji-Rong Wen}.} \bibinfo{year}{2023}\natexlab{}.
\newblock \showarticletitle{Large language models for information retrieval: A survey}.
\newblock \bibinfo{journal}{\emph{arXiv preprint arXiv:2308.07107}} (\bibinfo{year}{2023}).
\newblock


\end{thebibliography}

\onecolumn
\newpage
\appendix

\section{Prompt Library}

\subsection{Prompts on ALFWorld}

ALFWorld includes 6 different types of task, and we only present the prompt for the Put task here.

\subsubsection{Thought preparation}

We write thoughts for the same demonstration (\texttt{\$Demo} \texttt{1} and \texttt{\$Demo} \texttt{2}) as the first two in \emph{ReAct} \cite{yao2023react} and use them for thought preparation.

\begin{lstlisting}[basicstyle=\ttfamily\tiny,numbers=none,frame=lines,showspaces=false,breaklines=true,breakindent=0pt]
You are an agent to interact with a household to solve a task. You will be given a task where you need to put an (two) object(s) to a target either directly or after an operation. Each time you first think about your current situation, then output an action, and wait for next observation.
Here is your action space:
* go to target: Move to the target, and you will observe what is in/on the target or know it is closed or opened.
* open target: Open the target when it is closed, and you will observe what is in/on the target. Only cabinets, drawers, fridges, safes, and microwaves can be opened.
* take object from target: Take the object from the target when the object is in/on the target. You can take only one object at the same time.
* put object in/on target: Put an object you have taken/picked up in/on the target. You should go to the target in your last action. You can put no matter there are other objects in/on the target or not.
* clean object with target: Clean an object you have taken/picked up with the target. The target should be a sinkbasin. You should go to the target in your last action. You can clean no matter there are other objects in/on the target or not.
* heat object with target: Heat an object you have taken/picked up with the target. The target should be a microwave. You should go to the target in your last action. You can heat no matter there are other objects in/on the target or not.
* cool object with target: Cool an object you have taken/picked up with the target. The target should be a fridge. You should go to the target in your last action. You can cool no matter there are other objects in/on the target or not.
* use object: Use the object. The object should be a desklamp. You should be in/on a place where the object appears.
* look: Look around and see what you are facing. Only look when nothing happens.
Here are two examples.
$Demo 1
$Demo 2

Here is the task.
$Input
\end{lstlisting}

For the Put task we show part of \texttt{\$Demo} \texttt{1} here.

\begin{lstlisting}[basicstyle=\ttfamily\tiny,numbers=none,frame=lines,showspaces=false,breaklines=true,breakindent=0pt]
You are in the middle of a room. Looking quickly around you, you see a cabinet 4, a cabinet 3, a cabinet 2, a cabinet 1, a countertop 1, a garbagecan 1, a handtowelholder 2, a handtowelholder 1, a sinkbasin 2, a sinkbasin 1, a toilet 1, a toiletpaperhanger 1, and a towelholder 1.
Your task is to: put some spraybottle on toilet.
act: go to cabinet 1
think: I am now in/on: the middle of a room
Critical objects I have found:
None
Objects I have taken:
None
To solve the task, I need to find and take a sparybottle, then put it on toilet. First I need to find a spraybottle. A spraybottle is more likely to appear in cabinet (1-4), countertop (1), toilet (1), sinkbasin (1-2), garbagecan (1). I can check one by one, starting with cabinet 1.
obs: On the cabinet 1, you see a cloth 1, a soapbar 1, a soapbottle 1.
act: go to cabinet 2
think: I am now in/on: cabinet 1
Critical objects I have found:
None
Objects I have taken:
None
In cabinet 1, there is no spraybottle. Next, I need to check cabinet 2.
......
obs: On the toilet 1, you see a soapbottle 2.
act: put spraybottle 2 in/on toilet 1
think: I am now in/on: toilet 1
Critical objects I have found:
spraybottle 2 (cabinet 2)
Objects I have taken:
spraybottle 2
Now I am on the toilet 1. Next, I need to put spraybottle 2 in/on it.
obs: You put the spraybottle 2 in/on the toilet 1.
\end{lstlisting}

\subsubsection{Thought generation}

$~$

\begin{lstlisting}[basicstyle=\ttfamily\tiny,numbers=none,frame=lines,showspaces=false,breaklines=true,breakindent=0pt]
You are an agent to interact with a household to solve a task. You will be given a task where you need to put an (two) object(s) to a target either directly or after an operation. Each time you first think about your current situation, then output an action, and wait for next observation.
Here is your action space:
* go to target: Move to the target, and you will observe what is in/on the target or know it is closed or opened.
* open target: Open the target when it is closed, and you will observe what is in/on the target. Only cabinets, drawers, fridges, safes, and microwaves can be opened.
* take object from target: Take the object from the target when the object is in/on the target. You can take only one object at the same time.
* put object in/on target: Put an object you have taken/picked up in/on the target. You should go to the target in your last action. You can put no matter there are other objects in/on the target or not.
* clean object with target: Clean an object you have taken/picked up with the target. The target should be a sinkbasin. You should go to the target in your last action. You can clean no matter there are other objects in/on the target or not.
* heat object with target: Heat an object you have taken/picked up with the target. The target should be a microwave. You should go to the target in your last action. You can heat no matter there are other objects in/on the target or not.
* cool object with target: Cool an object you have taken/picked up with the target. The target should be a fridge. You should go to the target in your last action. You can cool no matter there are other objects in/on the target or not.
* use object: Use the object. The object should be a desklamp. You should be in/on a place where the object appears.
* look: Look around and see what you are facing. Only look when nothing happens.
When you cannot find some object, remember to check less possible places that you have not visited, or even impossible places like garbagecan.
Here are two examples.
$Demo 1
$Demo 2

Here is the task.
$Input
\end{lstlisting}

For the Put task we show part of a typical \texttt{\$Demo} here.

\begin{lstlisting}[basicstyle=\ttfamily\tiny,numbers=none,frame=lines,showspaces=false,breaklines=true,breakindent=0pt]
You are in the middle of a room. Looking quickly around you, you see a cabinet 1, a cabinet 2, a cabinet 3, a cabinet 4, a countertop 1, a garbagecan 1, a handtowelholder 1, a handtowelholder 2, a sinkbasin 1, a sinkbasin 2, a toilet 1, a toiletpaperhanger 1, and a towelholder 1.
Your task is to: put some soapbottle on toilet.
> think: I am now in/on: the middle of a room
Critical objects I have found:
None
Objects I have taken:
None
To solve the task, I need to find and take a soapbottle, then put it on the toilet. First, I need to find a soapbottle. A soapbottle is more likely to appear in cabinets (1-4), countertop (1), or sinkbasin (1-2). I will start by checking the garbagecan 1, although it's less likely to find a soapbottle there, it's still possible.
> act: go to garbagecan 1
On the garbagecan 1, you see a soapbottle 1.
> think: I am now in/on: garbagecan 1
Critical objects I have found:
soapbottle 1 (garbagecan 1)
Objects I have taken:
None
I have found a soapbottle (1) in the garbagecan 1. Next, I need to take it.
> act: take soapbottle 1 from garbagecan 1
You pick up the soapbottle 1 from the garbagecan 1.
> think: I am now in/on: garbagecan 1
Critical objects I have found:
soapbottle 1 (garbagecan 1)
Objects I have taken:
soapbottle 1
Now I have taken the soapbottle (1). Next, I need to go to and put it on the toilet 1.
> act: go to toilet 1
On the toilet 1, you see a cloth 1.
> think: I am now in/on: toilet 1
Critical objects I have found:
soapbottle 1 (garbagecan 1)
Objects I have taken:
soapbottle 1
Now I am at the toilet 1. Next, I need to put the soapbottle 1 on it.
> act: put soapbottle 1 in/on toilet 1
\end{lstlisting}

\subsubsection{Action prediction} For action prediction we use almost the same prompt as thought generation. For \emph{TRAD} we only use $B+1+F$ steps in each demonstration, and we add the following sentence to tell LLM the meaning of \emph{relative order mark}:

\begin{lstlisting}[basicstyle=\ttfamily\tiny,numbers=none,frame=lines,showspaces=false,breaklines=true,breakindent=0pt]
The mark [Step $i] in expert examples indicates a coarse relative position of expert demonstration steps to your situation. For example, [Step -1] means the last step, [Step 0] means the current step, and [Step 1] means the next step.
\end{lstlisting}

\subsection{Prompts on Mind2Web}

On Mind2Web we generally follow prompts in \emph{Synapse} \cite{zheng2023synapse}.

\subsubsection{Thought preparation}

$~$

\begin{lstlisting}[basicstyle=\ttfamily\tiny,numbers=none,frame=lines,showspaces=false,breaklines=true,breakindent=0pt]
You are a large language model trained to navigate the web. You will be given a task, an observation, and your previous actions, and each time you should output the next action and wait for the next observation. Here is the action space:
1. `CLICK [id]`: Click on an HTML element with its id.
2. `TYPE [id] [value]`: Type a string into the element with the id.
3. `SELECT [id] [value]`: Select a value for an HTML element by its id.
Now you are given some expert demonstrations and reasons for their actions, follow these examples and give your reason for the given action. Note that you should take all previous actions into reasoning, and not take the current action as what you have done.
$Demo 1
$Demo 2
$Demo 3
$Input
\end{lstlisting}

We show one demonstration here:

\begin{lstlisting}[basicstyle=\ttfamily\tiny,numbers=none,frame=lines,showspaces=false,breaklines=true,breakindent=0pt]
Task: Find JetBlue press releases for the year 2020
Trajectory:
obs: `<html> <jb-app> <jb-tab-panel tabpanel> <div> <div combobox> <jb-type-ahead-input> <label> From </label> <input id=908 text columbus port columbus intl apt, /> </jb-type-ahead-input> </div> <div combobox> <jb-type-ahead-input> <label> To </label> <input id=927 text /> </jb-type-ahead-input> </div> </div> </jb-tab-panel> <jb-footer contentinfo> <div> <a id=1554> Investor Relations <jb-icon img external link should open in /> </a> <a id=107> Press Room <jb-icon img external link should open in /> </a> </div> </jb-footer> <div> <textarea text /> <div> 250 characters remaining </div> <button id=1911> Submit </button> </div> </jb-app> </html>`
act: `CLICK [107]` ([link]  Press RoomExternal Link should open in a new windo... -> CLICK)
obs: `<html> <main main> <span> <a id=4386> View all releases </a> <div> <div> <a download> </a> <a id=5509 print> <span> Print Core Overhead Bins </span> </a> </div> <div> <a download> </a> <a id=6087 print> <span> Print Welcome Kiosk (JFK) </span> </a> </div> <div> <a download> </a> <a id=6614 print> <span> Print Core Overhead Bins </span> </a> </div> <div> <a download> </a> <a id=7192 print> <span> Print Welcome Kiosk (JFK) </span> </a> </div> </div> </span> </main> </html>`
act: `CLICK [4386]` ([link]  View All Releases \uedbe -> CLICK)
obs: `<html> <main main> <div> <div> <label id=8053> Select year: </label> <select id=7685> <option 2023 true> 2023 </option> <option 2022> 2022 </option> <option 2021> 2021 </option> <option 2020> 2020 </option> <option 2019> 2019 </option> <option 2018> 2018 </option> <option 2017> 2017 </option> <option 2016> 2016 </option> <option 2015> 2015 </option> <option 2014> 2014 </option> <option 2013> 2013 </option> <option 2012> 2012 </option> <option 2011> 2011 </option> <option 2010> 2010 </option> <option 2009> 2009 </option> <option 2008> 2008 </option> <option 2007> 2007 </option> <option 2006> 2006 </option> <option 2005> 2005 </option> <option 2004> 2004 </option> <option 2003> 2003 </option> <option 2002> 2002 </option> </select> </div> <div> <label> Category: </label> <select id=7691> <option true> All Latest News </option> <option press-release> Press Releases </option> <option articles> Articles </option> </select> </div> <div> <label> Search: </label> <input id=8124 text newssearch /> <button id=8126 submit> <span> Search </span> </button> </div> </div> </main> </html>`
act: `SELECT [7685] [2020]` ([combobox]  Select year: -> SELECT: 2020)
reason: I have to find:
JetBlue press releases
for the year 2020
Now I have done:
`CLICK [107]` ([link]  Press RoomExternal Link should open in a new windo... -> CLICK)
`CLICK [4386]` ([link]  View All Releases \uedbe -> CLICK) to show all releases
Therefore, next I have to:
`SELECT [7685] [2020]` ([combobox]  Select year: -> SELECT: 2020) due to the condition `for the year 2020`
\end{lstlisting}

The input are presented in the same format as demonstrations without human-written reasons.

\subsubsection{Thought generation}

$~$

\begin{lstlisting}[basicstyle=\ttfamily\tiny,numbers=none,frame=lines,showspaces=false,breaklines=true,breakindent=0pt]
You are a large language model trained to navigate the web. You will be given a task, an observation, and your previous actions. Each time you should output the next action and wait for the next observation. Here is the action space:
1. `CLICK [id]`: Click on an HTML element with its id.
2. `TYPE [id] [value]`: Type a string into the element with the id.
3. `SELECT [id] [value]`: Select a value for an HTML element by its id.
Now you are given some expert demonstrations, follow these examples and conduct reasoning about your situation.
$Demo 1
$Demo 2
$Demo 3
$Input
\end{lstlisting}

We show part of a typical \texttt{\$Demo} here.

\begin{lstlisting}[basicstyle=\ttfamily\tiny,numbers=none,frame=lines,showspaces=false,breaklines=true,breakindent=0pt]
Task: Find cheapest cars available at San Francisco Airport for a day.
Trajectory:
obs: `<html> <main> <div> <label id=132> Pick-up location </label> <input id=372 pick-up location /> </div> <div> <div> <a id=859> <span> San Francisco </span> </a> <a id=896> <span> San Francisco Airport </span> </a> </div> <button id=1137 button> <div> Airports </div> </button> </div> </main> </html>`
previous actions:

reason: I have to find:
the cheapest cars
available at San Francisco Airport
for a day
Now I have done:
None
Therefore, next I have to:
`CLICK [896]` ([link]  San Francisco Airport -> CLICK) to select the pick-up location as San Francisco Airport.
obs: `<html> <body> <a id=1885> <span> Car rental </span> </a> <div> <div> <label> Pick-up location </label> <input id=1994 pick-up location san francisco airport, us (sfo) /> </div> <div> <button id=2006 button tue, mar 28> <div> <div> Pick-up date </div> <div> Tue, Mar 28 </div> </div> </button> <button id=2131 button fri, mar 31> <div> <div> Drop-off date </div> <div> Fri, Mar 31 </div> </div> </button> <button id=2251 button> <div> Search </div> </button> </div> </div> </body> </html>`
previous actions:
`CLICK [896]` ([link]  San Francisco Airport -> CLICK)
reason: I have to find:
the cheapest cars
available at San Francisco Airport
for a day
Now I have done:
`CLICK [896]` ([link]  San Francisco Airport -> CLICK) due to the condition `at San Francisco Airport`
Therefore, next I have to:
`CLICK [2131]` ([div]  Fri, Mar 31 -> CLICK) to select the drop-off date as "Fri, Mar 31"
obs: `<html> <body> <a id=5677> <span> Car rental </span> </a> <div> <div> <label> Pick-up location </label> <input id=5786 pick-up location san francisco airport, us (sfo) /> </div> <div> <table grid> <td gridcell> <span id=5951 checkbox 29 march 2023> <span> 29 </span> </span> </td> </table> <div> <button id=6248 button fri, mar 31> <div> <div> Drop-off date </div> <div> Fri, Mar 31 </div> </div> </button> <select id=6264 dropoff-time> <option 00:00> Midnight </option> <option 00:30> 12:30 AM </option> <option 01:00> 1:00 AM </option> <option 01:30> 1:30 AM </option> <option 02:00> 2:00 AM </option> <option 02:30> 2:30 AM </option> <option 03:00> 3:00 AM </option> <option 03:30> 3:30 AM </option> <option 04:00> 4:00 AM </option> <option 04:30> 4:30 AM </option> <option 05:00> 5:00 AM </option> <option 05:30> 5:30 AM </option> <option 06:00> 6:00 AM </option> <option 06:30> 6:30 AM </option> <option 07:00> 7:00 AM </option> <option 07:30> 7:30 AM </option> <option 08:00> 8:00 AM </option> <option 08:30> 8:30 AM </option> <option 09:00> 9:00 AM </option> <option 09:30> 9:30 AM </option> <option 10:00 true> 10:00 AM </option> <option 10:30> 10:30 AM </option> <option 11:00> 11:00 AM </option> <option 11:30> 11:30 AM </option> <option 12:00> Noon </option> </select> </div> <button id=6369 button> <div> Search </div> </button> </div> </div> </body> </html>`
previous actions:
`CLICK [896]` ([link]  San Francisco Airport -> CLICK)
`CLICK [2131]` ([div]  Fri, Mar 31 -> CLICK)
reason: I have to Find cheapest cars available at San Francisco Airport for a day.
\end{lstlisting}

\subsubsection{Action prediction}

We use the same sentence as in ALFWorld to tell LLM about the \emph{relative order mark}.

\begin{lstlisting}[basicstyle=\ttfamily\tiny,numbers=none,frame=lines,showspaces=false,breaklines=true,breakindent=0pt]
You are a large language model trained to navigate the web. You will be given a task, an observation, and your previous actions. Each time you should output the next action and wait for the next observation. Here is the action space:
1. `CLICK [id]`: Click on an HTML element with its id.
2. `TYPE [id] [value]`: Type a string into the element with the id.
3. `SELECT [id] [value]`: Select a value for an HTML element by its id.
Now you are given some expert demonstrations, follow these demonstrations and make your decision.
The mark [Step $i] indicates a coarse relative position of expert demonstration steps to your situation. For example, [Step -1] means the last step, [Step 0] means the current step, and [Step 1] means the next step.
Note that you should take all previous actions into reasoning. In your output, the action should be quoted by a pair of '`'.
$Demo 1
$Demo 2
$Demo 3
$Input
\end{lstlisting}

We show the format of demonstrations here:

\begin{lstlisting}[basicstyle=\ttfamily\tiny,numbers=none,frame=lines,showspaces=false,breaklines=true,breakindent=0pt]
Task: Look for a job opening in sales in San Fransisco, and if found, apply for the job.
obs: `<html> <body> <div> <nav navigation> <ul menubar> <li> <button id=8372 menuitem> <span> Research </span> </button> <div menu> </div> </li> </ul> </nav> <nav cargurus corporate information navigation> <ul menu> <a id=8721 menuitem olink> Our Team </a> <a id=8015 menuitem olink> Careers </a> </ul> </nav> </div> <ul> <a id=9178 our team> Our Team </a> <a id=9208 careers> Careers </a> </ul> </body> </html>`
previous actions:
`CLICK [117]` ([link]  Our Team -> CLICK)
act: `CLICK [8015]` ([menuitem]  olink -> CLICK)
\end{lstlisting}

The input are presented in the same format as demonstrations, except that they have no ground-truth actions.

\newpage

\section{Full Experiment Results}

\subsection{The Effect of $F$ and $B$}\label{ap:abl-temporal-expansion}

\begin{figure}[htbp]
    \centering
    \begin{subfigure}[b]{0.24\textwidth}
        \includegraphics[width=\textwidth]{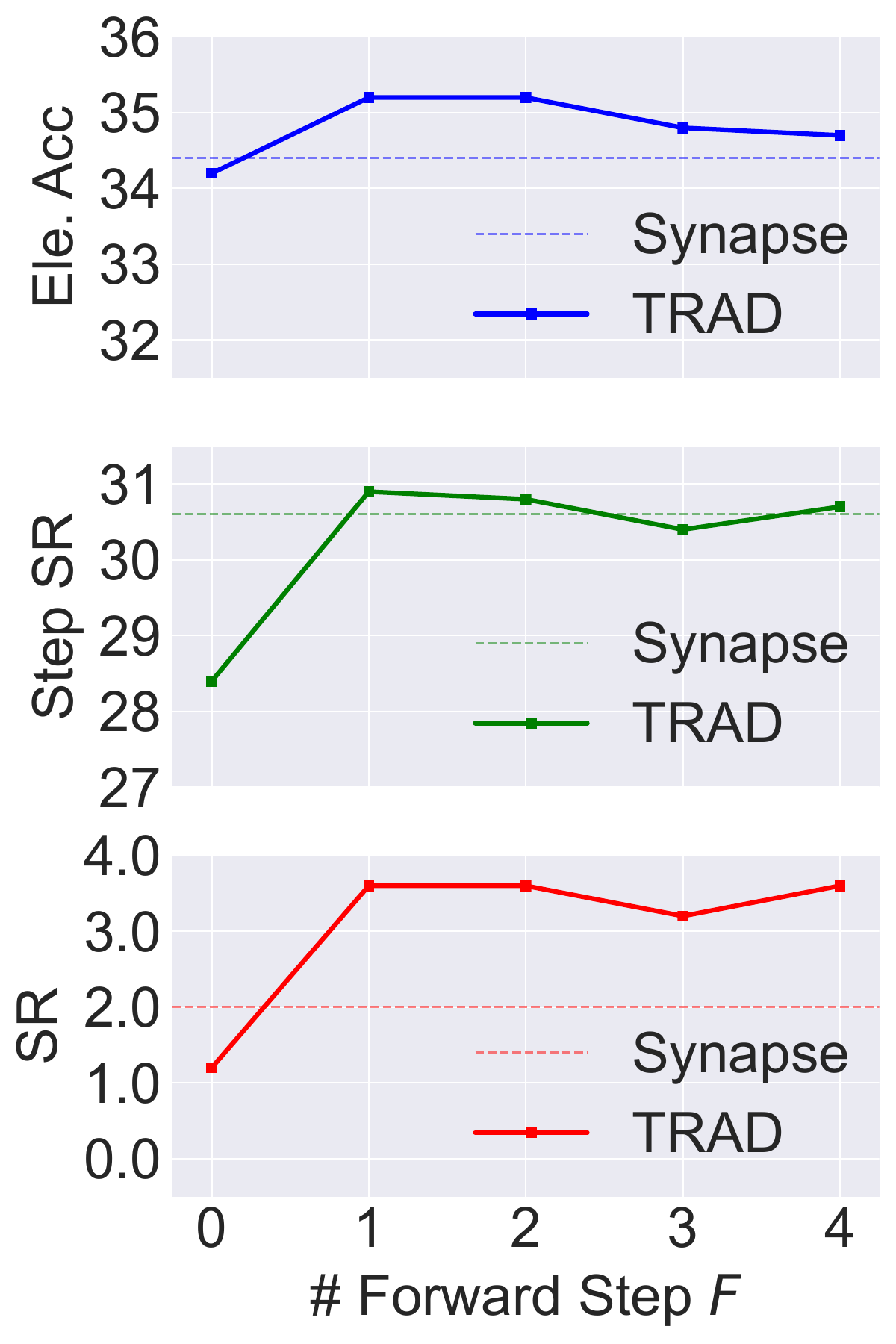}
        \vspace{-16pt}
        \caption{Cross-Task ($F$)}
        \label{fig:cross-task-forward}
    \end{subfigure}
    \begin{subfigure}[b]{0.24\textwidth}
        \includegraphics[width=\textwidth]{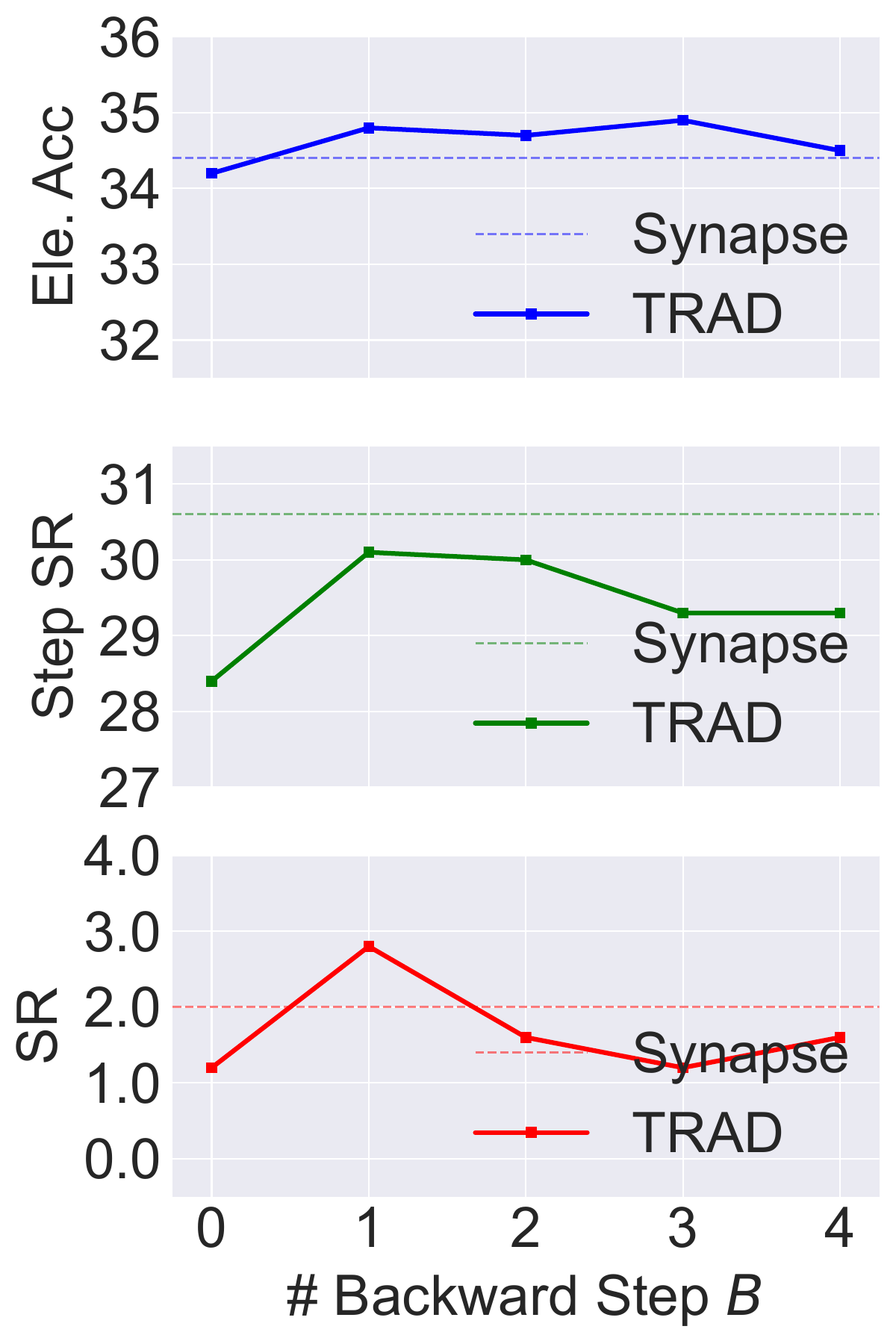}
        \vspace{-16pt}
        \caption{Cross-Task ($B$)}
        \label{fig:cross-task-backward}
    \end{subfigure}
    \begin{subfigure}[b]{0.24\textwidth}
        \includegraphics[width=\textwidth]{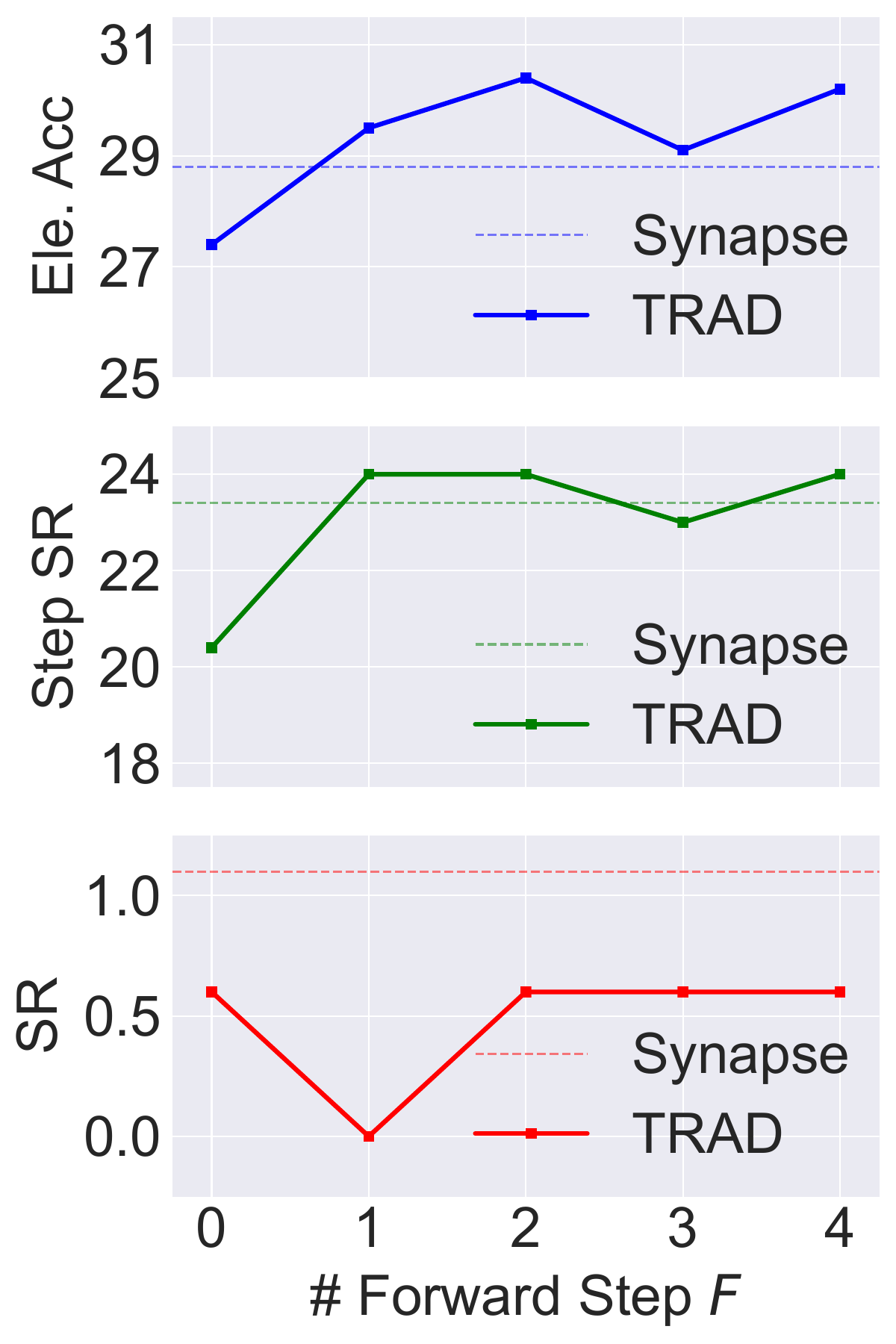}
        \vspace{-16pt}
        \caption{Cross-Website ($F$)}
        \label{fig:cross-website-forward}
    \end{subfigure}
    \begin{subfigure}[b]{0.24\textwidth}
        \includegraphics[width=\textwidth]{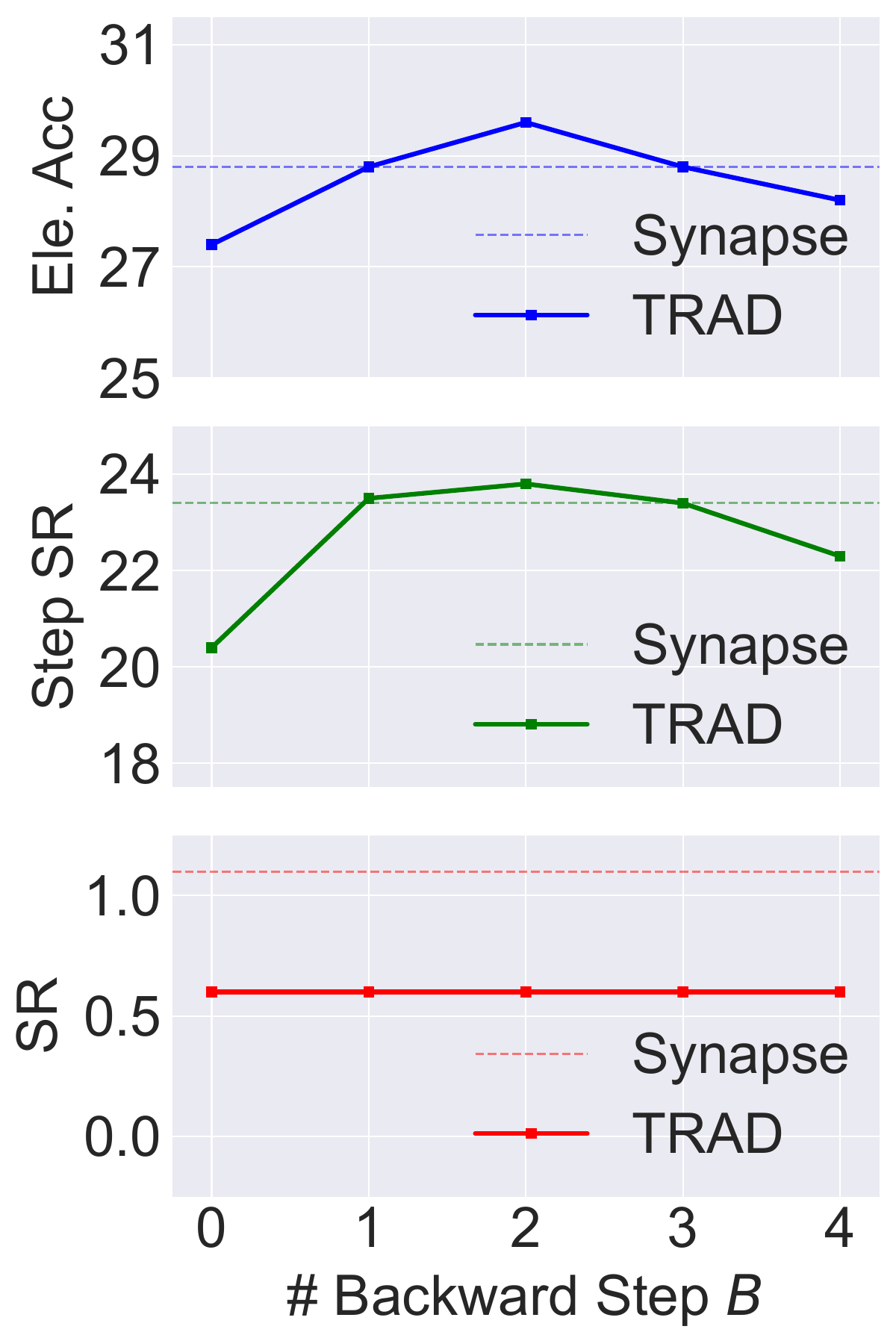}
        \vspace{-16pt}
        \caption{Cross-Website ($B$)}
        \label{fig:cross-website-backward}
    \end{subfigure}\\
    \begin{subfigure}[b]{0.24\textwidth}
        \includegraphics[width=\textwidth]{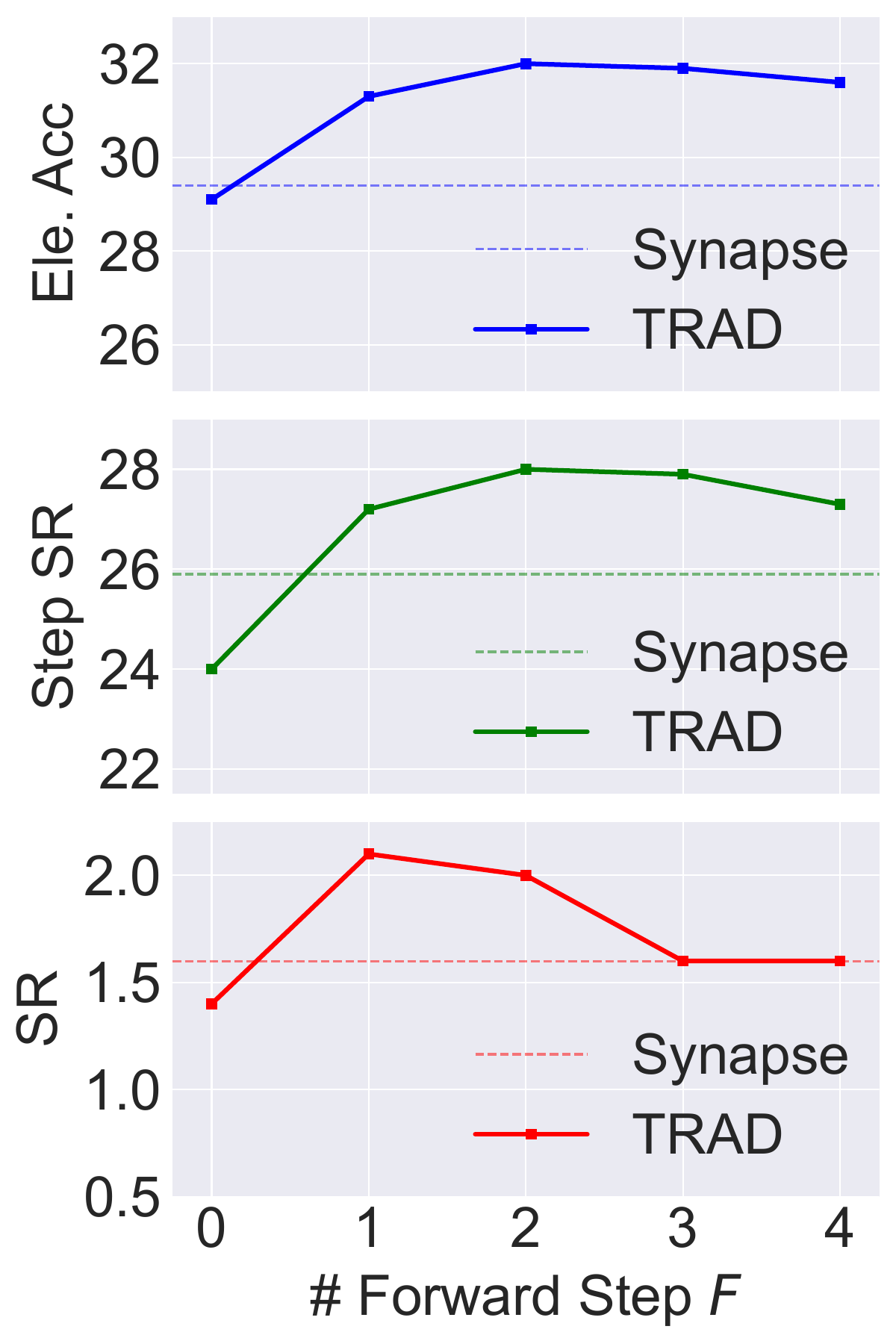}
        \vspace{-16pt}
        \caption{Cross-Domain ($F$)}
        \label{fig:cross-domain-forward}
    \end{subfigure}
    \begin{subfigure}[b]{0.24\textwidth}
        \includegraphics[width=\textwidth]{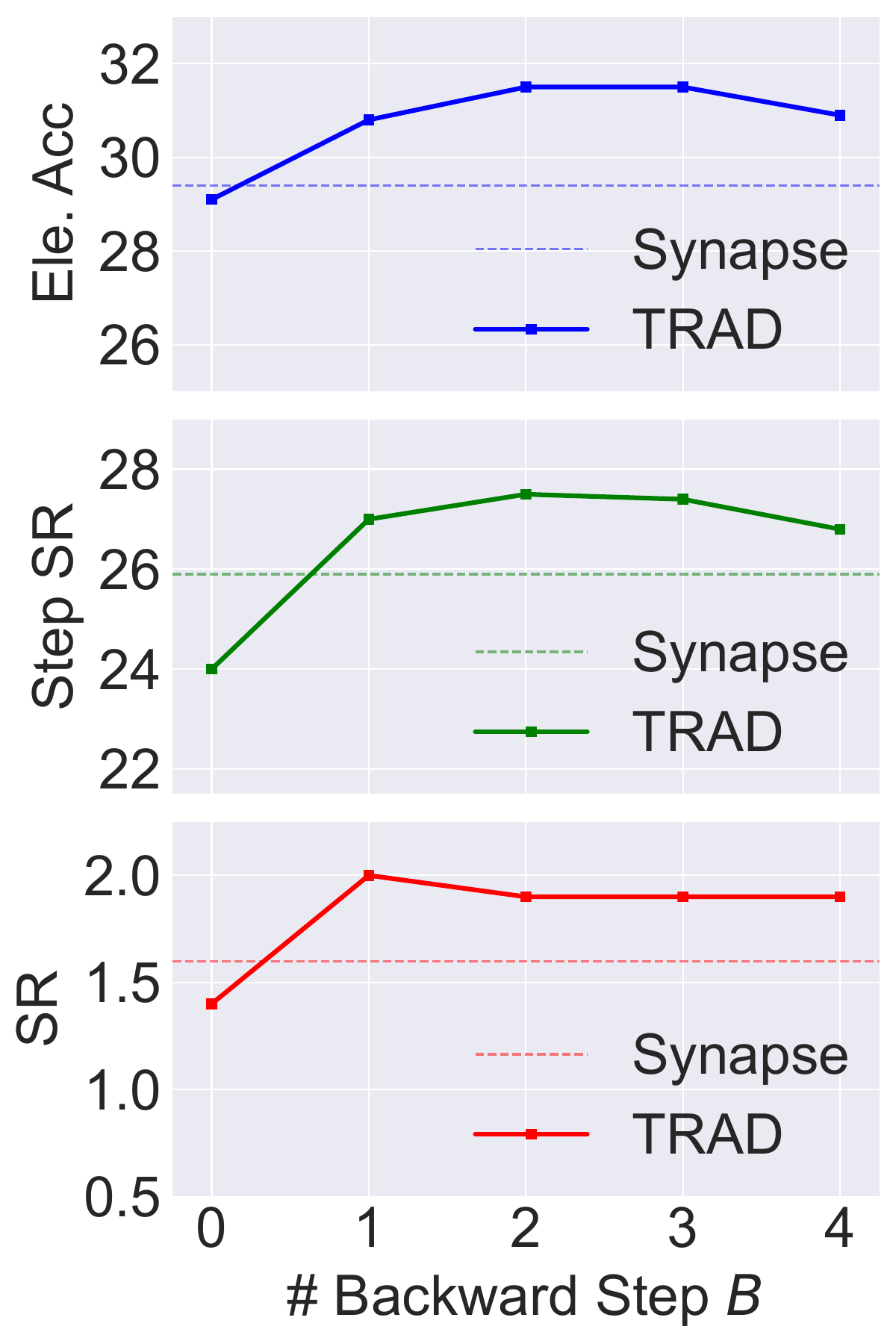}
        \vspace{-16pt}
        \caption{Cross-Domain ($B$)}
        \label{fig:cross-domain-backward}
    \end{subfigure}
    \begin{subfigure}[b]{0.24\textwidth}
        \includegraphics[width=\textwidth]{figure/ablation-expansion/average-forward.pdf}
        \vspace{-16pt}
        \caption{All ($F$)}
        \label{fig:average-forward}
    \end{subfigure}
    \begin{subfigure}[b]{0.24\textwidth}
        \includegraphics[width=\textwidth]{figure/ablation-expansion/average-backward.pdf}
        \vspace{-16pt}
        \caption{All ($B$)}
        \label{fig:average-backward}
    \end{subfigure}
    \vspace{-6pt}
    \caption{The effect of varying subsequent steps $F$ and previous steps $B$ on Mind2Web benchmark. Solid lines correspond to the performance metrics of \emph{TRAD} given different $F$ and $B$, and the dashed lines correspond to the \emph{Synapse} baseline. Forward expansion ($F>0$) generally provides more improvement than backward expansion ($B>0$) over no expansion ($F=B=0$) and the \emph{Synapse} baseline. $F$ or $B$ does not help more when they are sufficiently large.}
    \label{fig:exp-expansion-step-all}
\end{figure}

We list the results of varying subsequent step number $F$ and previous step number $B$ of temporal expansion on each subset and over all 3 subsets of the Mind2Web benchmark in \fig{fig:exp-expansion-step-all}.

\newpage

\subsection{The Effect of $K$}\label{ap:abl-k}

\begin{figure}[htbp]
    \centering
    \begin{subfigure}[b]{0.24\textwidth}
        \includegraphics[width=\textwidth]{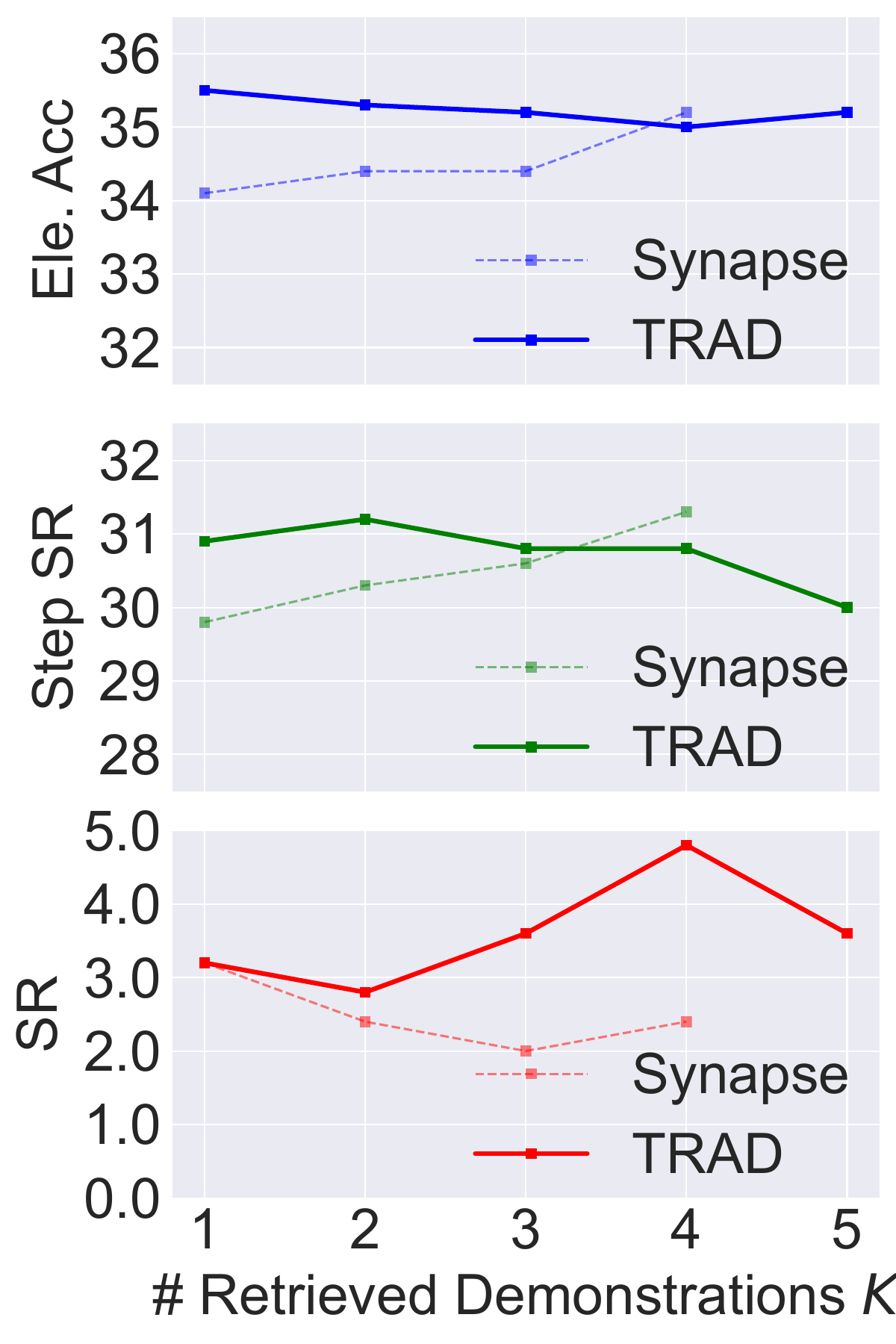}
        \vspace{-16pt}
        \caption{Cross-Task}
        \label{fig:cross-task-k}
    \end{subfigure}
    \begin{subfigure}[b]{0.24\textwidth}
        \includegraphics[width=\textwidth]{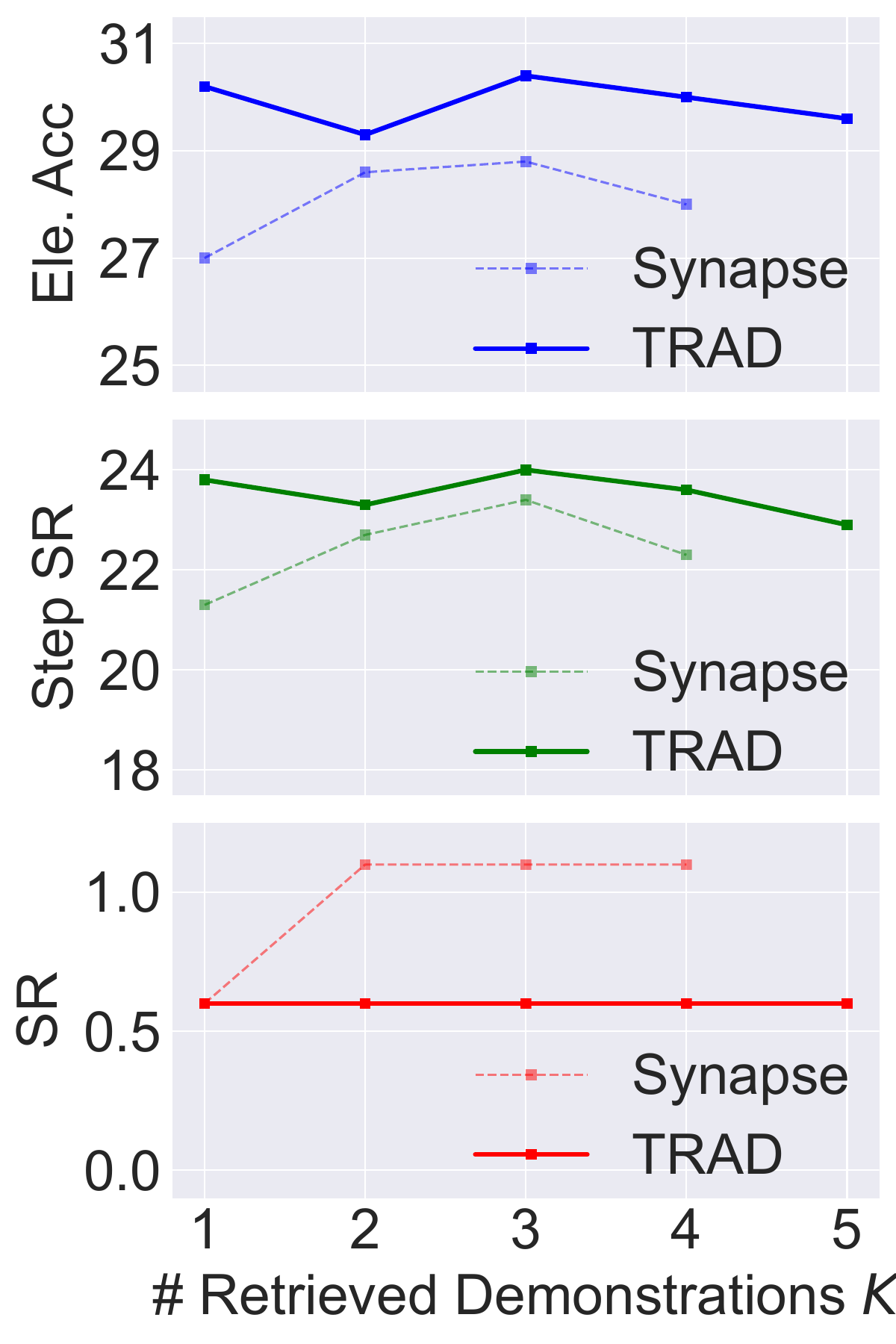}
        \vspace{-16pt}
        \caption{Cross-Website}
        \label{fig:cross-website-k}
    \end{subfigure}
    \begin{subfigure}[b]{0.24\textwidth}
        \includegraphics[width=\textwidth]{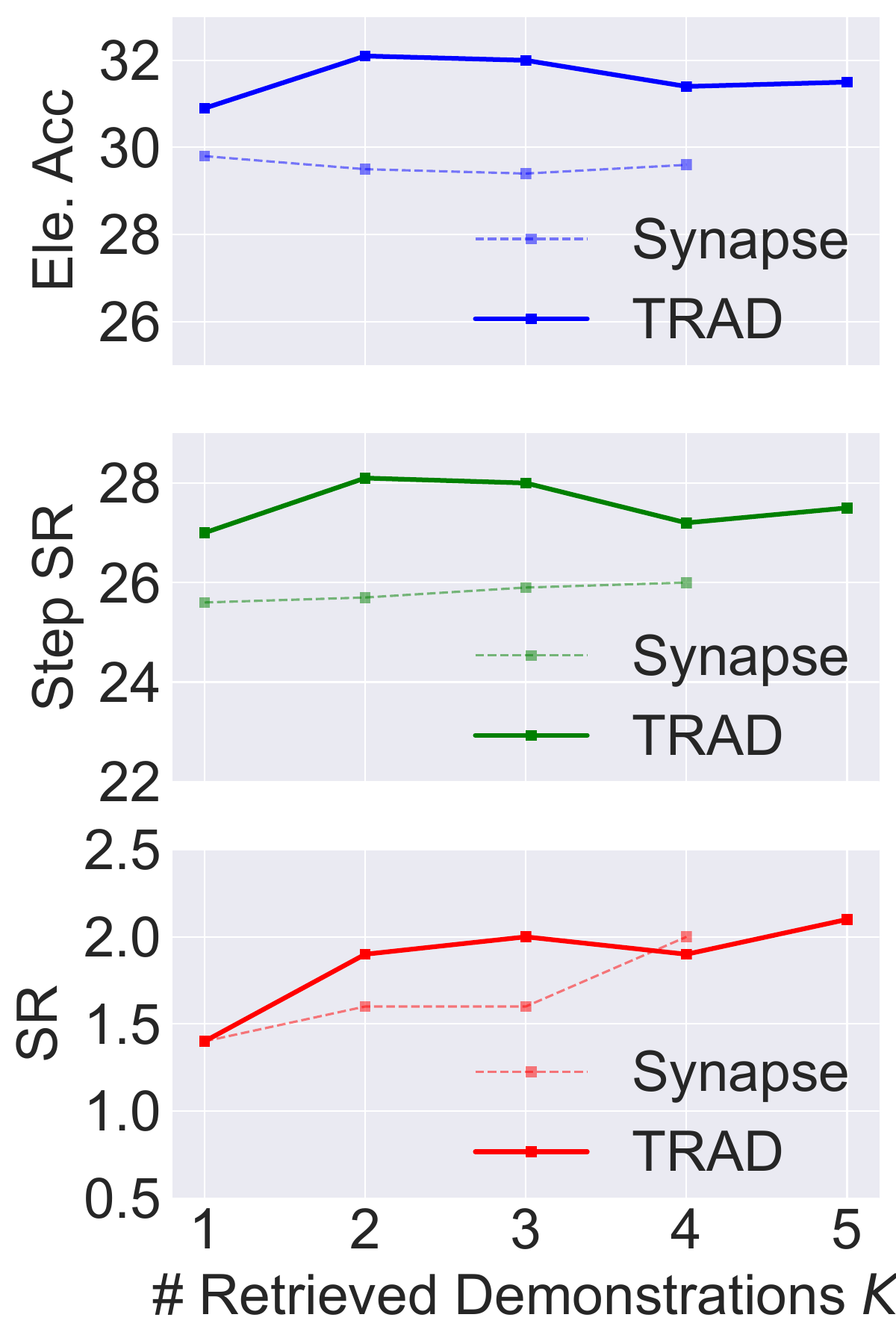}
        \vspace{-16pt}
        \caption{Cross-Domain}
        \label{fig:cross-domain-k}
    \end{subfigure}
    \begin{subfigure}[b]{0.24\textwidth}
        \includegraphics[width=\textwidth]{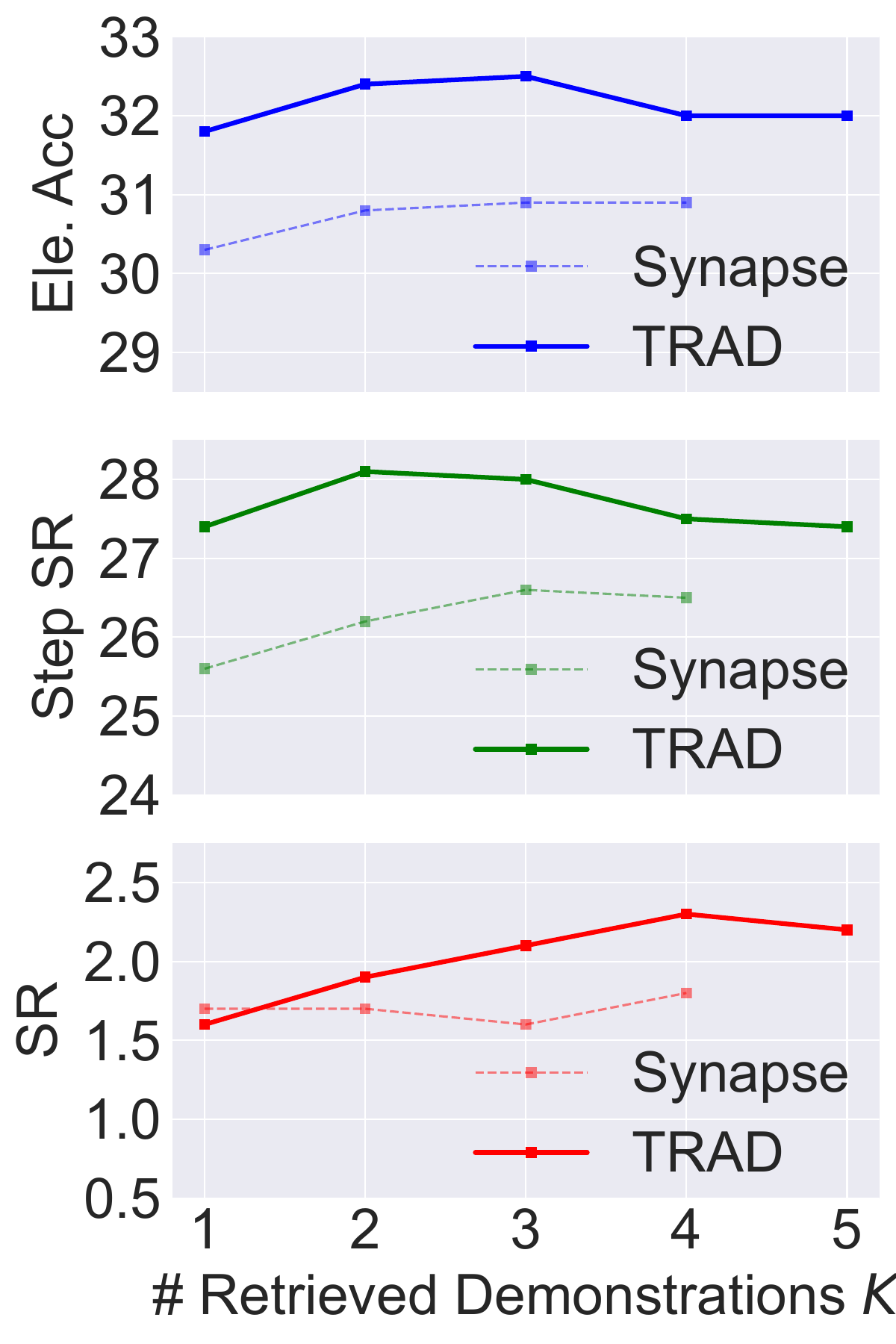}
        \vspace{-16pt}
        \caption{All}
        \label{fig:average-k}
    \end{subfigure}
    \vspace{-6pt}
    \caption{The effect of varying the number of retrieved demonstrations $K$ on Mind2Web benchmark. Solid lines correspond to the performance metrics of \emph{TRAD} given different $K$, and the dashed lines correspond to the \emph{Synapse} baseline. $K$ has a mild effect on the performance of \emph{TRAD} and \emph{Synapse}, and the advantage of \emph{TRAD} over \emph{Synapse} remains stable when $K$ varies.}
    \label{fig:exp-retrieval-k-all}
\end{figure}

We list the results of varying retrieval size $K$ on each subset and over all 3 subsets of the Mind2Web benchmark in \fig{fig:exp-retrieval-k-all}.

\end{document}